%% file: main.tex
\documentclass[10pt]{preprint}
\pdfinfoomitdate=1 \pdftrailerid{} \pdfsuppressptexinfo=-1 \hypersetup{pdfcreator={},pdfproducer={}}%

\usepackage{etoolbox}
\usepackage{arydshln}
\usetikzlibrary{arrows.meta,positioning,calc,decorations.pathreplacing,patterns}
\AtBeginEnvironment{tabular}{\fontfamily{ptm}\selectfont}
\AtBeginEnvironment{tikzpicture}{\fontfamily{ptm}\selectfont}
\DeclareCaptionFont{tabtimes}{\fontfamily{ptm}\selectfont}
\makeatletter
\let\flide@graphics\includegraphics
\newcommand{\plotgraphics}[2][]{\flide@graphics{#2}}
\AtBeginEnvironment{figure}{\let\includegraphics\plotgraphics}
\makeatother

\newcommand{\flideherefloat}[1]{%
  \expandafter\let\csname flide@#1\expandafter\endcsname\csname #1\endcsname
  \expandafter\let\csname flide@end#1\expandafter\endcsname\csname end#1\endcsname
  \renewenvironment{#1}[1][]{\csname flide@#1\endcsname[!tp]}{\csname flide@end#1\endcsname}}
\flideherefloat{table}\flideherefloat{table*}\flideherefloat{figure}\flideherefloat{figure*}
\titleformat{\paragraph}[runin]{\bfseries}{}{0pt}{}
\titlespacing*{\paragraph}{0pt}{0.6em}{0.5em}

\definecolor{tabaccent}{HTML}{2E8B8B}
\definecolor{stateD}{HTML}{1565C0}\definecolor{stateL}{HTML}{CEE6F3}
\definecolor{costD}{HTML}{C62828}\definecolor{costL}{HTML}{FBD2E0}
\definecolor{qaD}{HTML}{2E8B57}\definecolor{qaL}{HTML}{A7DDBC}
\definecolor{qbD}{HTML}{845B97}\definecolor{qbL}{HTML}{CABAE6}
\definecolor{qcD}{HTML}{D9822B}\definecolor{qcL}{HTML}{F8D9B5}
\definecolor{padD}{HTML}{6E7B83}\definecolor{padL}{HTML}{FFFFFF}
\definecolor{cardbg}{HTML}{F6F7F8}\definecolor{cardrule}{HTML}{D9DDE1}
\newsavebox{\cardbox}
\newenvironment{promptcard}[1]
  {\par\noindent\begin{lrbox}{\cardbox}\begin{minipage}{\dimexpr\linewidth-13pt\relax}\small\textbf{#1}\par\vspace{3pt}}
  {\end{minipage}\end{lrbox}{\setlength{\fboxsep}{6pt}\setlength{\fboxrule}{0.4pt}\fcolorbox{cardrule}{cardbg}{\usebox{\cardbox}}}\par}
\newcommand{\clause}[1]{\textcolor{tabaccent}{\textbf{#1}}}
\newcommand{\chattok}[1]{\textless\textbar #1\textbar\textgreater}
\newcommand{\tabsetup}{\fontsize{9}{10.5}\selectfont\renewcommand{\arraystretch}{1.0}\setlength{\tabcolsep}{5pt}}
\usepackage{pifont}
\usepackage{fontawesome5}
\newcommand{\cmark}{\textcolor{qaD}{\ding{51}}}
\newcommand{\xmark}{\textcolor{costD}{\ding{55}}}

\renewcommand{\brandlogofile}{assets/flide-logo}
\makeatletter
\providecommand\@LN[2]{}
\providecommand\@LN@col[1]{}
\makeatother

\begin{document}
\arrayrulecolor{black}

\setrunninghead{Flide: Accelerating Decision Inference with Exact State Forking}

\title{Flide: Accelerating Decision Inference\\ with Exact State Forking}

\setrunningauthor{Li et al.}

\author{Bobo Li\textsuperscript{1}} \author{Jiajia Song\textsuperscript{1}} \author{Hao Fei\textsuperscript{2}} \author{Wynne Hsu\textsuperscript{1}} \author{Mong Li Lee\textsuperscript{1}}
\affil{\textsuperscript{1}National University of Singapore \quad \textsuperscript{2}University of Oxford}
\affil{\normalfont\fontsize{9.5}{11}\selectfont\ttfamily \{libobo, jjiasong, dcshsuw, dcsleeml\}@nus.edu.sg \quad haofei7419@gmail.com}
\newlength{\linkwidth}
\settowidth{\linkwidth}{\normalfont\fontsize{9.5}{11}\selectfont \makebox[1.2em][l]{\faGithub}\textbf{Source Code:} \textrm{https://github.com/OmniJev/flide}}
\affil{\normalfont\fontsize{9.5}{11}\selectfont \makebox[\linkwidth][l]{\makebox[1.2em][l]{\faGlobe}\textbf{Project Page:} \href{https://flide.org}{\textrm{https://flide.org}}}}
\affil{\normalfont\fontsize{9.5}{11}\selectfont \makebox[\linkwidth][l]{\makebox[1.2em][l]{\faGithub}\textbf{Source Code:} \href{https://github.com/OmniJev/flide}{\textrm{https://github.com/OmniJev/flide}}}}

\maketitle

\begin{abstract}
Decision models such as Jev have emerged as a highly promising paradigm for LLM agent systems, providing lightweight and reliable categorical reasoning for rapid System One execution.
However, existing generation-oriented inference engines fail to support these workloads effectively, suffering from severe prefix recomputation overheads or distorted output probabilities.
To address this challenge, we propose Flide, an efficient and exact inference engine that decouples the logical tree structure of a decision request from its physical execution within a single flattened token stream.
Through ForkAttention, Flide navigates request dependencies using GPU-resident indices across both standard attention and recurrent Gated DeltaNet layers, guaranteeing in exact arithmetic that every query branch evaluates identically to an isolated forward pass.
By expressing branch topologies directly in GPU-resident metadata, Flide accommodates diverse request mixtures with a compact set of pre-recorded execution graphs while enabling expanding agent histories to resume from checkpoints saved in earlier passes.
Extensive evaluations on public hybrid and decision models demonstrate that Flide answers queries up to 9.5 times faster than vLLM and SGLang while staying closest to a float32 reference in total variation distance.
By unifying physical execution throughput with mathematical exactness, Flide establishes a principled foundation for scalable, latency-critical decision intelligence in agent systems.
\end{abstract}


\makeatletter
\begingroup
\renewenvironment{figure*}[1][]{\par\vspace{0.8em}\noindent\begin{minipage}{\linewidth}\centering\def\@captype{figure}}{\end{minipage}\par\vspace{0.4em}}
\input{figures/teaser}
\endgroup
\makeatother

\section{Introduction}
\label{sec:intro}

Large language models increasingly serve as core decision engines within complex systems for tasks such as routing requests \citep{ong2024routellm}, evaluating responses \citep{zheng2023judging}, and deciding the next steps of agents \citep{yao2023react}.
Generating text to make categorical choices wastes computational resources on unnecessary tokens and increases latency, leading to the emergence of Jev \citep{typesafe2026jev} and other specialized decision models \citep{startlux2026decision,openjev2026} for these tasks.
These models typically process extensive context, such as a lengthy conversation or an agent trajectory, to answer several questions about it, outputting probability distributions over multiple discrete choices.
When deployed on every routed request and agent action, with strict thresholds acting on their probabilities \citep{decisionindex2026}, these confidence-sensitive models call for an inference engine that is both fast and precise.

However, current inference engines severely constrain the application of decision models since they are optimized for text generation, where the fundamental unit of work is a single linear sequence.
When processing decisions, frameworks like vLLM \citep{vllm2026systemone} and SGLang \citep{sglang2026systemone} treat every question as an independent prompt, rendering, tokenizing and scheduling the shared context once for every question.
The latest vLLM computes the shared context of a request once before its questions \citep{vllm2026pr61192}, yet every question still runs as a sequence of its own and resumes from the prefix cache.
On hybrid architectures this cache restores recurrent states only at block boundaries \citep{pan2025marconi,jing2026radixflow}, and every question and every agent step thus computes the tokens past the last boundary again.
Alternatively, multi-item scoring \citep{sglang2026mis} appends all questions to a single context with a delimiter token before each.
Every question then reads this extra token, which changes its hidden states and corrupts its output probabilities compared to isolated prompts \citep{sglang2026misdecisions}.
Furthermore, under dynamic batching, the order of floating-point sums changes with the batch \citep{he2025nondeterminism}, which can yield fluctuating decisions for the identical request under varying server loads.
All of these issues severely compromise the reliability and efficiency of decision model serving.

The root cause is a mismatch between the logical structure of a decision request and the physical structure in which current engines execute it.
Logically, the questions of a request are separate branches of a tree whose trunk is the shared state, and each branch must see only the trunk and itself.
Physically, these branches should run together over a single copy of the trunk, which speeds up inference by computing the trunk only once and letting many short branches fill the GPU in one pass.
Generation engines tie the physical layout to the logical one, running every question as a separate sequence that computes or reloads the trunk again.
Multi-item scoring ties them the other way and does run the branches together.
Yet it writes the tree into the sequence itself, and its delimiter tokens enter the input of every branch and, as Figure~\ref{fig:teaser} illustrates, shift the probabilities and flip decisions.
Closing this gap means running the branches together while each still computes exactly what it would compute alone, so the physical stream must keep apart the branches it places side by side.
An attention layer can keep them apart with a mask, yet the Gated DeltaNet layers \citep{yang2025gdn} of hybrid models such as Qwen3.5 \citep{qwen2026qwen35} fold every earlier token into one recurrent state, through which a branch would inherit its siblings.
Separating the branches there requires each of them to restart from the exact state of its trunk, inside the same pass that computes the trunk.

To address this, we propose \textbf{Flide} (Flat-Graph Inference for Decision Models), an efficient and exact inference engine that separates the logical structure of decision requests from their physical execution on the GPU.
Flide lays a round of requests out as one flat token stream and records their dependencies in GPU-resident metadata.
With ForkAttention, the attention and Gated DeltaNet layers follow these dependencies, and in exact arithmetic every question computes what it would compute alone.
Since the request structure is encoded in data, a small set of recorded execution graphs can serve different request compositions.
The same state-forking mechanism also supports growing agent histories, allowing later requests to resume from checkpoints saved during earlier inference passes.
In addition, an optional batch-invariant mode fixes floating-point reduction orders to keep answers bitwise identical across server loads.

We evaluate Flide against state-of-the-art inference engines on workloads ranging from one question per passage to articles and agent trajectories with many questions.
Experimental results show that Flide answers up to 9.5 times faster than vLLM and SGLang on public hybrid models and leads on dedicated decision models.
Its throughput leads at every concurrency level whenever a shared state carries several questions.
Meanwhile, as multi-item scoring shifts probabilities to gain throughput, Flide yields outputs closest to the float32 reference distributions in total variation distance.
These results establish that separating the logical structure of decision requests from their physical execution lets an inference engine be exact and fast at once.

Our contributions are threefold.
\begin{itemize}
\item We trace both the repeated computation and the shifted answers of current engines to one root cause, the coupling between the logical structure of a decision request and its physical execution.
\item We propose Flide, which keeps each request as a tree in GPU metadata while running every round as one flat stream, and forks every state exactly with ForkAttention.
\item On public hybrid and decision models, Flide answers up to an order of magnitude faster than vLLM and SGLang while staying closest to the float32 reference in total variation.
\end{itemize}
\section{Related Work}
\label{sec:related}

\paragraph{Prefix Sharing and State Branching.}
Modern serving systems reduce redundant computation by caching key-value representations of reusable prompt prefixes \citep{kwon2023vllm,zheng2024sglang,zheng2024batchllm}.
Specialized execution kernels retrieve these cached representations during decoding via cascade attention \citep{ye2025flashinfer,juravsky2024hydragen,ye2024chunkattention,zhu2024relayattention} or sequence-tree traversals \citep{yao2025deft}.
For hybrid architectures containing recurrent states \citep{gu2023mamba,yang2025gdn}, existing frameworks either constrain state materialization to coarse block boundaries \citep{pan2025marconi,vllm2026prefix,jing2026radixflow} or require predefined checkpoint positions \citep{shirokikh2026sparse}.
Speculative decoding verifies token trees with tree attention \citep{miao2024specinfer,cai2024medusa}, and its hybrid variants propagate recurrent states along the branches \citep{wu2025stree,ghantasala2026treewy,ma2026lumotree}.
Domain-specific frontends fork requests into isolated sequences that query a shared prefix cache \citep{zheng2024sglang}.
However, all of these mechanisms mandate a strict temporal separation where shared prefixes must complete execution and populate memory buffers before downstream queries can consume them.
Flide eliminates this decoupled caching phase by evaluating both the shared context and every branching question within a single forward pass, dynamically propagating both attention and Gated DeltaNet states across a flattened token sequence.

\paragraph{Decision Model Serving and Execution Exactness.}
Decision-making architectures operate under strict latency budgets and rely on well-calibrated categorical distributions \citep{typesafe2026jev,onejev2026,openjev2026,vllmsr2026d3}.
Existing engines handle these single-token workloads either by processing questions as isolated sequences \citep{du2025prefillonly,vllm2026systemone,sglang2026systemone}, which repeats prefix calculations, or by concatenating candidate queries with delimiter tokens \citep{sglang2026mis}, which invalidates isolated query semantics and requires CPU-managed segmentation that breaks CUDA graph capture.
Standard graph capture strategies mitigate prefill dispatch overhead by padding requests to static shapes \citep{sglang2026cudagraph}, but still treat the individual sequence as the primitive unit of execution.
Even when the prefix is computed once beforehand \citep{vllm2026pr61192}, every question resumes from the cache as a sequence of its own.
Other specialized variants incur latency penalties by scanning block masks \citep{chouhan2026encode,antunes2026ndm}, packing spoken input streams into eager passes \citep{jin2026duplexjev}, or evaluating candidate trees across segmented passes \citep{ma2026fastjev}.
Compounding these systemic overheads, LLM serving engines let co-scheduled requests change the outputs of one another, whereas database systems share work across concurrent queries while keeping each query isolated \citep{sellis1988mqo}.
Prefix cache substitutions create observable discrepancies compared to isolated evaluation \citep{patodiya2026samerequest}, and dynamic reduction orders cause floating-point nondeterminism \citep{he2025nondeterminism}.
Existing batch-invariant kernels slow serving by 34 percent on Qwen3-8B in SGLang \citep{sglang2025deterministic}, and the batch-invariant mode of vLLM cannot execute recurrent Gated DeltaNet layers \citep{vllm2026gdninvariant}.
Flide resolves these scheduling and numerical conflicts simultaneously by executing multi-branch rounds through GPU-resident metadata.
This approach preserves full CUDA graph replay across dynamic workloads while every answer equals its standalone computation in exact arithmetic.
As summarized in Table~\ref{tab:engines}, Flide is the only one of these engines that concurrently achieves single-pass execution, unpolluted query inputs, arbitrary hybrid state branching, CUDA graph replaying, and an optional bitwise-invariant mode under dynamic loads.

\input{tables/engines}

\section{Method}
\label{sec:method}

\input{figures/method}

As illustrated in Figure~\ref{fig:method}, Flide keeps the tree of each request as its logical structure and runs every round of requests as one flat token stream on the GPU.

\subsection{Problem Formulation}
\label{sec:problem}

A decision model receives a state and several questions about it in one request, as in the System One API \citep{typesafe2026jev} that vLLM \citep{vllm2026systemone} and SGLang \citep{sglang2026systemone} serve.
A decision request $r = (s, q_1, \dots, q_n)$ comprises a state $s$ of $P$ tokens and $n$ questions, where question $q_i$ has $|q_i|$ tokens, as illustrated in Figure~\ref{fig:method}(a).
Flide renders the prompt for every question exactly as it would appear if asked in isolation.
It defines $s$ as the longest sequence of leading tokens that all such prompts share before the text of their questions begins.
Consequently, the concatenation $s \oplus q_i$ matches the standalone prompt for $q_i$ token for token.
Let $h_i$ denote the final hidden state at the last token of $q_i$ when the model processes $s \oplus q_i$.
Each question $q_i$ comes with a set of options $O_i$, and the engine computes the probability of each option as
\begin{equation}
p_i(o) = \frac{\exp(w_o^\top h_i)}{\sum_{o' \in O_i} \exp(w_{o'}^\top h_i)}, \qquad o \in O_i,
\label{eq:answer}
\end{equation}
where $w_o$ is the row of the output head corresponding to option $o$.
This row belongs to a label token in a chat model or to a trained slot in a decision model.
In exact arithmetic, $p_i$ depends only on $s$ and $q_i$, remaining independent of other concurrent questions or requests.

Naively processing $s \oplus q_i$ as a separate sequence for every question requires $nP + \sum_i |q_i|$ token computations, which redundantly evaluates the shared state $n$ times.
Computing the state and each question exactly once reduces this cost to $P + \sum_i |q_i|$.
While prefix caching approximates this efficiency, it still forces every question to execute as an independent sequence that resumes from the cache.
To reach this count, Flide computes the shared state and all of its questions jointly in a single forward pass, which raises three challenges once every answer must stay exact.
\begin{itemize}
\item \textbf{First}, every question must keep its exact standalone computation, which rules out any change to its input such as a delimiter before each question.
Hybrid models make this harder, since their Gated DeltaNet layers \citep{yang2025gdn} compress all prior tokens into a single recurrent state and leave no attention matrix to mask.
\item \textbf{Second}, rounds mixing requests of arbitrary numbers and lengths must run with minimal GPU launch overhead, given that a short decision request takes only a few milliseconds of GPU time.
CUDA graphs remove these launches but fix the shape of every launch, whereas the tree of a round changes from one round to the next.
\item \textbf{Third}, when a state grows incrementally over the successive steps of an agent, the engine must reuse the unchanged part of the previous state and compute only its short mutable tail and the newly appended tokens.
\end{itemize}

\subsection{ForkAttention}
\label{sec:tree}

To resolve the first challenge, ForkAttention structures each decision request as a computational tree where the shared state forms a common trunk and the questions fork as independent branches.
Evaluating the shared trunk once allows every question to inherit the prefix representation without redundant computation or prompt modifications.
Pointwise operations, including linear projections, normalization layers, and multilayer perceptrons, remain indifferent to this tree structure because they process each token in complete isolation.
State forking therefore requires intervention only in the token-mixing layers, which must grant questions access to the shared trunk while isolating sibling branches from each other.
Flide maps this tree onto the GPU by flattening an entire round of requests into a single continuous token stream without padding.
State tokens appear first, followed by their corresponding question branches,
\begin{equation}
\bigl[\, s^{(1)} \mid s^{(2)} \mid \cdots \mid q^{(1)}_1 \mid \cdots \mid q^{(1)}_{n_1} \mid q^{(2)}_1 \mid \cdots \bigr],
\label{eq:stream}
\end{equation}
where $s^{(j)}$ denotes the state trunk of the $j$-th request and $q^{(j)}_c$ denotes its $c$-th question branch, as illustrated in Figure~\ref{fig:method}(b).
Every token in this flattened stream carries metadata indicating its request index $j$, its branch index $c$ (with $c = 0$ designating the trunk), and its sequential position $t$.
The tree thus remains the logical structure of every request, the stream is only its physical layout, and these indices map every token from one to the other.
Positional indices for each question continue from $P$, and the rotary embedding of any question token therefore matches the one it receives in the standalone sequence $s \oplus q_i$.
If the state incorporates images, vision encoder representations replace image placeholder tokens once per state.
The state retains the native multimodal rotary positions of the model, enabling each question to resume from the position immediately following the state boundary.

\subsubsection{Full-Attention Layers}
A query token $x$ attends to a key token $y$ when the following conditions hold,
\begin{equation}
j_y = j_x, \qquad c_y \in \{0, c_x\}, \qquad t_y \le t_x,
\label{eq:visible}
\end{equation}
which guarantees that the key belongs to the same request, resides either in the shared state or within the query's specific question, and does not appear later in the sequence.
Each question thus attends only to its trunk and itself as in $s \oplus q_i$, similar to the tree attention that speculative decoding applies to candidate tokens \citep{miao2024specinfer}.
A sliding-window layer of width $W$, as in Gemma 4 \citep{gemmateam2026gemma4}, adds $t_x - t_y < W$, which selects the same keys as in $s \oplus q_i$ because the positions agree.

\subsubsection{Gated DeltaNet Layers}
Hybrid models replace attention with Gated DeltaNet \citep{yang2025gdn} in most of their layers.
Such a layer folds every token into a single matrix-valued recurrent state $\mathbf{S}_t$ of fixed size by the gated delta rule,
\begin{equation}
\begin{aligned}
\mathbf{S}_t &= \alpha_t \mathbf{S}_{t-1} \bigl(\mathbf{I} - \eta_t \mathbf{k}_t \mathbf{k}_t^\top\bigr) + \eta_t \mathbf{v}_t \mathbf{k}_t^\top, \\
\mathbf{o}_t &= \mathbf{S}_t \mathbf{q}_t,
\end{aligned}
\label{eq:gdn}
\end{equation}
where $\mathbf{q}_t$, $\mathbf{k}_t$ and $\mathbf{v}_t$ are the query, key and value vectors of token $t$, and the decay $\alpha_t$ and write strength $\eta_t$ are computed from the token.
A short causal convolution in front of the rule further mixes each token with its few predecessors.
Since $\mathbf{S}_t$ already contains every earlier token, a mask cannot separate two questions that share a sequence.
Each question therefore has to continue the recurrence from $\mathbf{S}_P$, the recurrent state at the last token of $s$.
Flide therefore runs each Gated DeltaNet layer in two calls over the stream of Equation~\ref{eq:stream}.
The first call processes the state segments of all requests and outputs $\mathbf{S}_P$ and the convolution window of each request.
The second call processes the question segments of all requests at once, and each question starts from the $\mathbf{S}_P$ of its own request through an index without copying any state.
The convolution reads the predecessors of each token through the same kind of index, which lets the first tokens of a question see the last tokens of its state.
Every question thus continues from the exact state its standalone input would reach, at any position and with every token computed once, as illustrated in Figure~\ref{fig:method}(b).

\subsubsection{Exactness}
In a causal model, where the state never attends to a question, the visibility rule of Equation~\ref{eq:visible} and the two calls of every Gated DeltaNet layer together give each question the computation of its standalone prompt.

\begin{proposition}
\label{prop:exact}
For a causal model in exact arithmetic, the hidden state computed by Flide at any token of question $q_i$ is identical to the hidden state at that token when the model processes $s \oplus q_i$ in isolation, regardless of other requests and questions in the same round.
\end{proposition}

The proof proceeds by induction over the layers and is provided in Appendix~\ref{app:proof}.
In other words, the result that the tree of a request defines holds under any flat layout of the round.
In floating point, the differences stay at the rounding level, which the tests of Appendix~\ref{app:tests} check on every change of the engine.

\subsection{Flat-Graph Execution}
\label{sec:graphs}

Flat-graph execution resolves the second challenge by turning the forks of a round into GPU work with negligible launch overhead, as illustrated in Figure~\ref{fig:method}(c).
A short decision request on a small model takes only a few milliseconds of GPU time, which launching several thousand kernels one by one would dominate.
Although CUDA graphs eliminate these sequential launches, a captured graph fixes the shape of every kernel launch, whereas a server encounters rounds containing requests of highly variable numbers and lengths.
To reconcile this, Flide keeps the tree of a round purely as GPU data.
The shape of every launch then depends on the length of the flat stream alone, and a small collection of captured graphs accommodates any mix of requests and questions.

\subsubsection{GPU-Resident Metadata}
All variables determined by the round composition, including segment lengths, positional indices, key ranges, and state indices, reside in compact device buffers populated before graph replay, from which every kernel derives its workload.
The attention kernel fetches keys from two sources, the shared state of the request and the query's specific question, and skips the states of other requests interleaved in the stream.
The kernel constructs its tile table directly on the GPU, loads the fully visible key blocks without applying a mask, and restricts masking exclusively to partially visible blocks.
The Gated DeltaNet layers iterate over a list of fixed-length chunks populated by each round, with any unused entries beyond the active chunks returning immediately.
Consequently, altering the mix of requests only modifies the buffer contents while preserving the exact launch configuration of the captured graph.

\subsubsection{Graphs over Stream Lengths}
Building on length-bucketed CUDA graph capture \citep{sglang2026cudagraph}, Flide allows rounds with different decision-tree structures to reuse the same captured graph.
At server initialization, Flide captures a family of flat graphs indexed by stream capacity $T$.
Each graph also reserves fixed capacities for requests, question segments, Gated DeltaNet chunks, and readout rows.
The actual request boundaries and question dependencies are supplied through GPU-resident metadata, allowing different request mixes to share the same execution graph.
For each round, Flide selects the smallest graph whose buffers can accommodate both its flattened stream and associated metadata.
Tokens are packed without padding between requests; padding occurs only at the tail, where dense layers still perform computation but no valid token accesses the padded positions.
If a round exceeds the largest graph's capacity, Flide splits it along request boundaries into multiple replays.
An individual request exceeding this capacity runs through the same flat-stream execution without CUDA graph replay.

\subsubsection{Selective Readout}
The output head of Equation~\ref{eq:answer} reads only the last token of every question, so in the final layer every token still contributes its keys and values while the rest of the layer, its MLP included, runs only at these rows.
An index buffer of fixed size in every graph holds the rows, which skips most of the final-layer work without changing any output in exact arithmetic.

\subsection{Incremental State Forking}
\label{sec:grow}

Successive steps of an agent trajectory often share most of their state, and therefore most of the trunk of their decision trees.
Flide exploits this overlap by checkpointing the state of a request at an internal position $m$.
A later request can resume from the checkpoint when its tokens up to $m$ match.
Each checkpoint stores the float32 recurrent state and convolution window of every Gated DeltaNet layer at $m$, together with the keys and values of the full-attention layers before $m$.
This allows a resumed request to compute the same function as a fresh one in exact arithmetic.
Flide places $m$ slightly before the end of a state so that tokens likely to change at the next step, such as those at the end of a conversational turn, fall after the checkpoint.

Flide generates the checkpoint during the same forward pass that computes the request's answers.
To achieve this, Flide divides the state segments of the stream into two levels at their checkpoints.
Each Gated DeltaNet layer processes the two levels sequentially, saving the final states of the first level as checkpoints and passing them to the second through the same state-indexing mechanism used for question forking.
For full attention, cached keys and values precede newly computed ones in a buffer representing each state's history, allowing resumed states to be processed without occupying a contiguous span of the flat stream.
At each agent step, Flide thus computes the new tokens and a short uncached tail only once for all questions, saves a checkpoint for the next step, and forks the questions from the completed state, as illustrated in Figure~\ref{fig:method}(d).

\subsection{Bitwise Determinism}
\label{sec:bits}

Proposition~\ref{prop:exact} makes the result of the tree independent of the flat layout in exact arithmetic, yet in floating point a matrix product over more rows may pick another kernel, tile shape or reduction split and reorder the sums of an output element \citep{he2025nondeterminism}.
Flide therefore offers an optional mode, in the spirit of batch-invariant inference \citep{he2025nondeterminism}, in which no reduction depends on the composition of the round.
Every matrix product sums each output element in a fixed order whatever the number of rows, and every question gets attention tiles and Gated DeltaNet chunks of its own, which start its reductions exactly at the fork.
A question asked alone also runs as its state and one branch in this mode, and every answer computed from a full state therefore keeps the bits of the question asked alone under any load on the same hardware and software.

\section{Experiments}

\label{sec:exp}

\input{tables/workloads}

\subsection{Settings}
\label{sec:settings}

\paragraph{Workloads.}
We build six workloads from four public datasets, RACE \citep{lai2017race}, QuALITY \citep{pang2022quality}, MultiRC \citep{khashabi2018multirc} and the Decision Index \citep{decisionindex2026}.
Two of them send one question per state, with RACE passages and with QuALITY articles cut to 3,600 tokens.
The other four send every question of a state in one request, as the datasets come, from four questions per RACE passage to 48 per MultiRC paragraph on average, as Table~\ref{tab:workloads} lists and Appendix~\ref{app:workloads} details.

\paragraph{Baselines.}
We compare against vLLM \citep{kwon2023vllm} and SGLang \citep{zheng2024sglang} through their decision endpoints, against SGLang's multi-item scoring \citep{sglang2026mis}, MIS in the tables, and on the dedicated decision models against vllm-jev \citep{modeio2026vllmjev} and its RadixFlow scheduler \citep{jing2026radixflow}.
vLLM, SGLang and Flide serve the instruct models of Qwen3.5 \citep{qwen2026qwen35} from 0.8B to 27B and of Gemma 4 \citep{gemmateam2026gemma4} from E2B to 31B, multi-item scoring the Qwen3.5 models and vllm-jev Open-Jev-2B and Open-Jev-9B \citep{openjev2026}, all in bfloat16 with CUDA graphs on, except multi-item scoring, which turns them off.
Flide and vLLM read the same prompts, while SGLang and vllm-jev render their own templates, and Appendix~\ref{app:engines} lists the versions and settings of every engine.

\paragraph{Metrics.}
Lat. is the median latency in milliseconds of a request from a single client, and C8, C32 and C64 are the requests per second under 8, 32 and 64 closed-loop clients.
Against a float32 forward pass of the same model, Agr. is the share of questions with the same top answer, TV the mean total variation distance in points, P99 the 99th percentile of the largest change in one option's probability, and Flip the share of decisions that a threshold of 0.5 takes differently.
Acc. is the share of correct answers, and Stab. the share of questions whose top answer at 64 clients equals the one at a single client.

\paragraph{Setup.}
All measurements run on one NVIDIA H200 GPU, with the engines started one after another and every prefix cache emptied before each level.
Every row is sent four times, twice for DI and the full QuALITY articles, and each copy's state starts with its own tag, which keeps any request from reusing the state of another.

\subsection{Main Results}
\label{sec:main}

\input{tables/main}

\begin{figure*}[t]
\centering
\includegraphics[width=\linewidth]{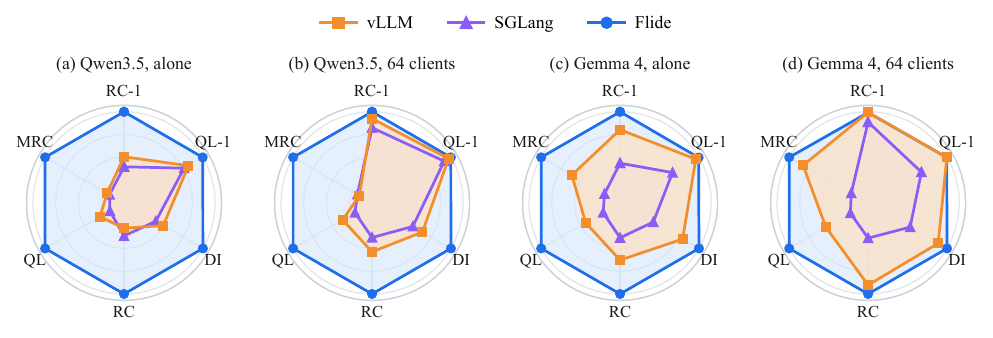}
\caption{Flide leads on every workload where a state carries several questions. Speed of each engine over the fastest one, geometric mean over the four sizes. RC, QL and MRC stand for RACE, QuALITY and MultiRC, and -1 marks one question per state.}
\label{fig:radar}
\end{figure*}

Table~\ref{tab:main} compares the three engines on Qwen3.5 and Gemma 4, while Figure~\ref{fig:radar} summarizes their relative speeds across workloads, from which we draw three findings.

First, Flide is faster even with a single question per state.
On Qwen3.5, it cuts the latency of a lone request by up to 3.2 times on the smallest model.
The gain comes from one flat-graph replay per round, against the heavier front ends and schedulers of generation engines.
Under load, where model computation dominates, Flide matches the best baseline.

Second, the advantage grows substantially when multiple questions share a state.
Flide leads at every model size and concurrency level, achieving up to 9.5 times lower latency and 8.5 times higher throughput than the best baseline.
The largest gains occur on QuALITY and MultiRC, which carry the most questions per state, since ForkAttention computes the shared state once for all of them while conventional engines repeatedly compute or restore it.
Multi-item scoring also shares the state but introduces delimiter tokens and runs without CUDA graphs.
Flide achieves higher throughput while preserving each question's standalone computation in exact arithmetic.

Finally, Flide also maintains its advantage on full-attention models.
Even against vLLM's prefix cache, which already computes the shared state of Gemma 4 once, Flide cuts the latency of multi-question requests by 1.1 to 2.9 times.
These results highlight the efficiency of executing an entire request in one forward pass, beyond the computational savings provided by prefix caching.

\subsection{Answer Fidelity}
\label{sec:fidelity}

\input{tables/fidelity}

Table~\ref{tab:fidelity} reads the answers of the Qwen3.5 models, and averaged over the four sizes, Flide's top answer agrees with the float32 reference most often on every workload.
Flide's accuracy stays within 1.2 points of vLLM's in every cell, so it gains its speed at the same accuracy, and SGLang's differs more only on MultiRC, where it words yes-or-no questions its own way.
Batching changes the order of floating-point sums, so every engine changes some answers between one and 64 clients, Flide at most 0.2 percent, SGLang up to 1.7 and vLLM up to 3.2 percent.
In the mode of Section~\ref{sec:bits}, Flide returns the same bits at any load, as Appendix~\ref{app:analysis} shows.

\subsection{In-depth Analysis}
\label{sec:analysis}

Further analyses, from identical bits and the time one request spends in each part of an engine to the start-up of a server, follow in Appendix~\ref{app:analysis}.

\input{tables/ablation}

\paragraph{Ablation.}
Table~\ref{tab:ablation} removes individual components of Flide on Qwen3.5-4B to isolate their performance contributions.
Without state forks, each question executes as an independent sequence, causing the latency of a multi-question QuALITY article to increase by more than an order of magnitude.
Without CUDA graphs, a solitary request launches thousands of sequential kernels, yielding a severe latency penalty on short inputs.
Without flat graphs, every request necessitates a distinct replay, stripping a round of short RACE requests of over one third of its throughput.
Finally, selective readout delivers further throughput gains by avoiding redundant token projections across all workloads.
Each component thus removes a distinct cost, from repeated state computation to kernel launches, graph replays and final-layer work, and together they give Flide its end-to-end speed.

\input{tables/vllmjev}
\paragraph{Dedicated decision models.}
Table~\ref{tab:vllmjev} evaluates Open-Jev, the exact architecture vllm-jev and its RadixFlow scheduler were designed to serve.
Flide outperforms both baselines across all configurations, achieving roughly threefold speedups on short inputs and maintaining an advantage of up to 1.7 times on long prompts where the forward pass dominates execution time.
This demonstrates that a generalized engine operating over the logical request tree outpaces heavily specialized systems.

\begin{figure}[t]
\centering
\includegraphics[width=\linewidth]{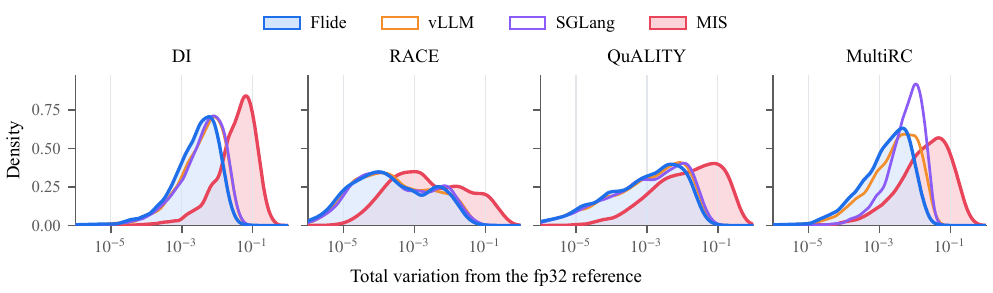}
\caption{Multi-item scoring shifts the whole distribution. Total variation of each answer from the float32 reference on Qwen3.5-4B, every question of a state in one request.}
\label{fig:tv}
\end{figure}

\paragraph{The cost of a delimiter.}
Multi-item scoring shares context by inserting a delimiter token before each question, and Figure~\ref{fig:tv} illustrates the resulting penalty on Qwen3.5-4B.
The median outputs of this delimiter approach diverge from the float32 reference roughly twelve times as much as those of Flide, whereas the curves of Flide, vLLM, and SGLang overlap.
Across all sizes and workloads, the delimiter method flips up to 29 percent of categorical decisions, compared to a maximum of 2.3 percent for Flide, as detailed in Table~\ref{tab:exact} of Appendix~\ref{app:analysis}.
Properly sharing the internal state thus keeps Flide closest to the float32 reference in total variation, confirming that the shifted answers of multi-item scoring stem directly from its input modifications.

\begin{figure}[t]
\centering
\includegraphics[width=\linewidth]{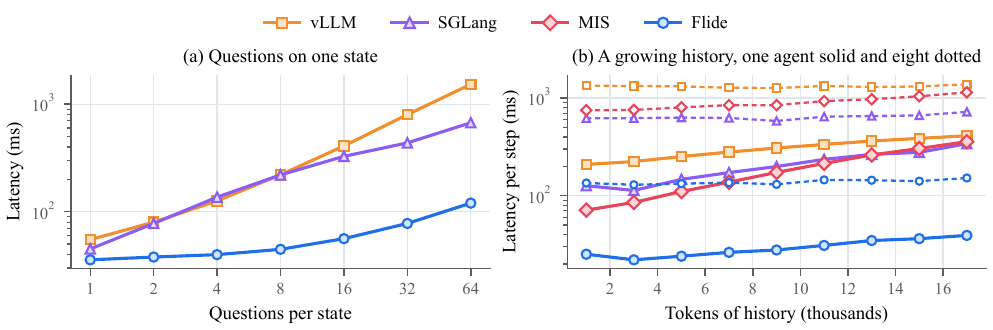}
\caption{Flide adds little latency per question and per step. Latency on Qwen3.5-4B as the questions on one state and the history of an agent grow.}
\label{fig:scaling}
\end{figure}

\paragraph{Questions per state.}
Figure~\ref{fig:scaling}a isolates a state of roughly two thousand tokens from a QuALITY article and evaluates performance across a growing number of questions.
Because vLLM and SGLang process each question as an independent sequence, their latencies increase drastically as the questions grow.
Flide computes the shared state exactly once and only appends the tokens specific to each branch.
With 64 questions on the state, Flide operates 12.7 times faster than vLLM and maintains a substantial lead over SGLang.
This follows the cost analysis of Section~\ref{sec:problem}, as the stream grows only by the tokens of each question, while separate sequences repeat the whole state for every question.

\input{tables/traj}

\paragraph{States that grow.}
We replay dozens of SWE-agent trajectories \citep{yang2024sweagent} from a public dataset \citep{nebius2024swetraj}, truncating them after their initial steps to yield over a thousand total states, and pose multiple questions against the complete history at every turn.
As Table~\ref{tab:traj} and Figure~\ref{fig:scaling}b illustrate, the marginal latency cost per thousand tokens on Flide stays around one millisecond.
In contrast, vLLM and SGLang suffer an order of magnitude higher latency penalty by running each question independently, while multi-item scoring degrades severely by recomputing the entire history at every step.
Regardless of the agent count, Flide ultimately completes 4.4 to 7.2 times the steps per second of the strongest baseline.

\section{Conclusion}
\label{sec:conclusion}

In this paper, we presented Flide, an efficient and exact inference engine for decision models that decouples the logical tree structure of each request from its physical execution within a single flattened token stream.
By leveraging ForkAttention, Flide forks intermediate states precisely across both attention and recurrent layers, replaying arbitrary execution rounds through compact pre-recorded graphs while extending agent histories directly from checkpoints saved in earlier passes.
This architecture resolves the fundamental conflict between hierarchical request logic and sequential token execution, eliminating both the redundant prefix computations and the delimiter tokens that shift the answers of existing serving engines.
Extensive evaluations across hybrid and full-attention architectures, specialized decision benchmarks, and real-world agent trajectories demonstrate that Flide achieves substantial speedups over current state-of-the-art frameworks while staying closest to the float32 reference distributions on Qwen3.5.
By unifying high-throughput parallel execution with strict mathematical exactness, Flide establishes a principled foundation for scalable, latency-critical decision intelligence in next-generation LLM agent systems.

\bibliographystyle{plainnat}
\bibliography{custom}

\appendix
\section{Proof of Proposition~\ref{prop:exact}}
\label{app:proof}

Fix a request with state $s$ of $P$ tokens and a question $q_i$.
Every token of $s \oplus q_i$ enters the stream with the token id and position it has in the standalone sequence, and norms, projections, the MLP and the rotary embedding act on one token at a time.
A full-attention layer lets each token of $q_i$ see exactly the tokens before it in $s \oplus q_i$ by Equation~\ref{eq:visible}, and each Gated DeltaNet layer continues $q_i$ from the recurrent state $\mathbf{S}_P$ and the convolution window that $s$ reaches alone.
Equal inputs to a layer therefore give equal outputs, and induction over the layers proves the proposition, since no step involves another request or question of the round.
A checkpoint restores the values that a fresh pass reaches at its position, and selective readout only drops rows that no read uses, so both keep this equality.

\section{Engine Versions and Launch Settings}
\label{app:engines}

\input{tables/launch}
Table~\ref{tab:launch} lists the engines, and every cell of vLLM reports the faster of its two builds, while the analyses and the comparisons on growing states, repeated states and images run the build of 14 October.
SGLang and vllm-jev use their own prompt formats, and multi-item scoring receives the state as the query and every question as an item, as the token ids that Flide and vLLM read.
The token-id rows of Table~\ref{tab:split} send the same ids to the completion endpoint of vLLM and to the score endpoint of SGLang without multi-item scoring.
For the images of GQA, vLLM and SGLang serve their chat endpoints with one request per question, thinking off and one output token, and a label outside the 20 most likely tokens counts as zero.
The float32 reference runs with TF32 disabled, a threshold of 0.5 acts when the probability of yes or of the top option reaches it, and the client sends 32 warm-up requests before each level.

\section{Workload Construction}
\label{app:workloads}

Table~\ref{tab:workloads} lists the six workloads as the engines receive them, built from public sets by scripts in the release with a fixed seed.
Its token counts come from the tokenizer of Qwen3.5, and those of the Gemma 4 tokenizers stay within 8 percent of them.
RACE samples 256 questions of the middle-school test set of RACE with their passages \citep{lai2017race}, and RACE all samples 256 passages of the same set with all of their questions.
QuALITY samples 256 questions of the validation set of QuALITY \citep{pang2022quality} and cuts each article at a line boundary so that the article, the question and its longest option fit 3,600 tokens with room for the template, inside the 4,096-token context of the Open-Jev models, while QuALITY all keeps all 115 validation articles at full length with all of their questions.
The comparison with vllm-jev runs its two Open-Jev models on RACE and on this cut QuALITY.
MultiRC \citep{khashabi2018multirc} turns every candidate answer of a question into the yes-or-no question \emph{question} Is ``\emph{answer}'' a correct answer?, adding whole questions to a paragraph while it holds at most 64.
DI samples 760 requests of the public set of the Decision Index \citep{decisionindex2026}, and 81 of them hold a choice question with more than 26 options, more letters than the alphabet holds, so 679 requests with 3,510 choice and 125 yes-or-no questions remain, 2,422 of them with a gold answer.
Every engine reads at most 26 options per choice question, 10 levels per score question and 64 questions per request, and every other workload satisfies these limits as built.
In the cold setting each copy of a row begins its state with a line naming the copy and the row, so no request shares a prompt prefix with another, while the questions of one request still share their state on every engine.

\begin{figure}[t]
\begin{promptcard}{Prompt of one question, in the wording of vLLM}
\ttfamily\footnotesize
\chattok{im\_start}user\\
\textit{state}\\[4pt]
Question: \textit{question}\\
\clause{options}\\
\clause{reply}\chattok{im\_end}\\
\chattok{im\_start}assistant\\
\textless think\textgreater\\[4pt]
\textless/think\textgreater\\[6pt]
\normalfont\footnotesize
\clause{options} are \texttt{A: }\textit{name}\texttt{ - }\textit{description} per option of a choice question, \texttt{yes} and \texttt{no} for a yes-or-no question, and \texttt{0: }\textit{level} per level of a score question.
\clause{reply} is \texttt{Answer with the letter of one option only.}, \texttt{Answer with yes or no only.} or \texttt{Answer with the number of one level only.}
\end{promptcard}
\caption{The prompt of one question. The answer is the next token over the option labels.}
\label{fig:prompt}
\end{figure}

Flide and vLLM render a question as vLLM's decision endpoint renders it for chat models, one user turn with the state, a blank line and the question, followed by the opening of the assistant turn with thinking turned off, as shown in Figure~\ref{fig:prompt}.
The answer is the distribution of the next token over the labels of the options, renormalized over those labels.
In Flide the part of the prompt before the question is the state of the stream and every question with the rest of the prompt is a branch, so its tokens are exactly those of vLLM's prompt.
SGLang writes structured states as compact JSON and opens a yes-or-no question with \texttt{Is the following true?} followed by the question, listing only the outcomes that carry a description, and the reference for SGLang is computed in this wording.

\section{Full Results}
\label{app:full}

\input{tables/full_q4b}
\input{tables/full_ge4b}

\begin{figure}[t]
\centering
\includegraphics[width=\linewidth]{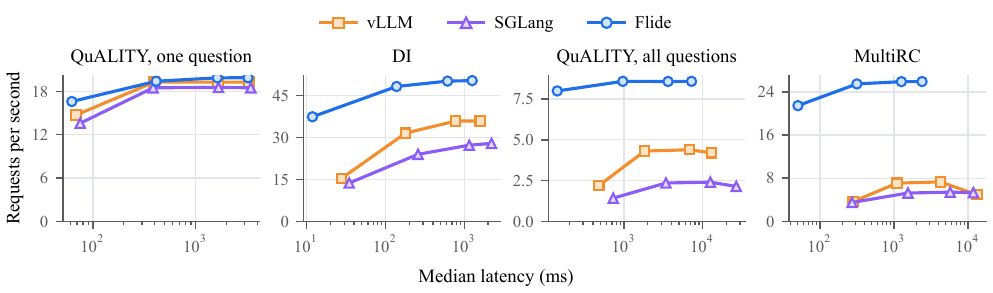}
\caption{Flide serves more requests at a lower latency. Requests per second against median latency for 1, 8, 32 and 64 clients on Qwen3.5-4B.}
\label{fig:load}
\end{figure}

\begin{figure}[t]
\centering
\includegraphics[width=\linewidth]{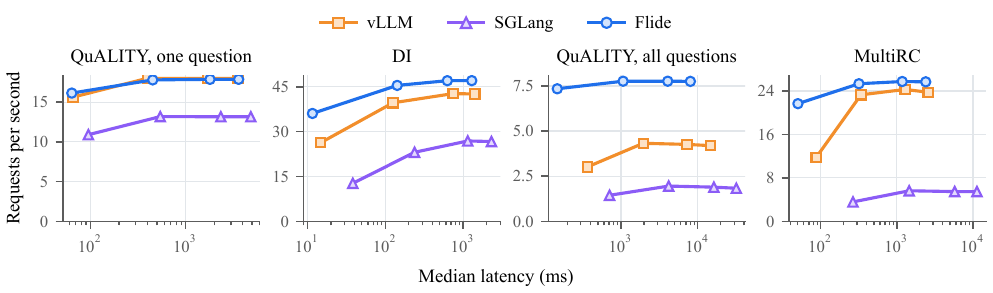}
\caption{Flide serves more requests at a lower latency on Gemma 4. Requests per second against median latency for 1, 8, 32 and 64 clients on Gemma 4 E4B.}
\label{fig:load_gemma}
\end{figure}

Figures~\ref{fig:load} and~\ref{fig:load_gemma} follow Qwen3.5-4B and Gemma 4 E4B from one to 64 clients.
Tables~\ref{tab:fullq4b} and~\ref{tab:fullge4b} list every level of each engine on every workload for Qwen3.5-4B and Gemma 4 E4B, where multi-item scoring serves only the Qwen3.5 workloads with every question of a state in one request, with the median and 99th percentile latency and the throughput.
With more than one question per request, Flide keeps the lowest 99th percentile latency at every number of clients on both models, which extends Table~\ref{tab:main} to the tail.

\section{Further Analysis}
\label{app:analysis}

\input{tables/exact}

\input{tables/det}

\paragraph{A growing state.}
Figure~\ref{fig:scaling}b follows the step latency of Section~\ref{sec:analysis} on Qwen3.5-4B over the full range of history lengths, with one agent and with eight.
Every engine keeps its cache on except multi-item scoring, whose launch settings turn its radix cache off, and each agent sends its next step once the last one is answered.

\paragraph{Identical bits.}
With the batch-invariant mode of Section~\ref{sec:bits}, every answer in Table~\ref{tab:det} keeps its bits at any load on all six models and six workloads.
On each Qwen3.5 model, every forked answer also has the bits of its question asked alone, on all 626 questions of DI, RACE, MultiRC and QuALITY.
The invariant mode costs at most 10 percent of the throughput at 64 clients, against the 34 percent that the batch-invariant kernels of SGLang lose on Qwen3-8B \citep{sglang2025deterministic}.

\input{tables/split}

\paragraph{Where the time goes.}
In Table~\ref{tab:split}, which takes one request on the 0.8B model apart, vLLM fed token ids computes the same tokens on QuALITY and MultiRC as behind its endpoint.
The 187~ms it loses on a QuALITY article and the 57~ms on a MultiRC paragraph thus go to rendering and tokenizing the prompt of every question.
Below their front ends vLLM and SGLang still compute 1.5 to 6.5 times the tokens of Flide, since every question carries its own copy of the state and a prefix cache restores only part of it.
SGLang's multi-item scoring computes the tokens of Flide to within two percent, one delimiter per question, yet at every size it takes longer for a request alone, up to 8.0 times on the 0.8B model, and serves 1.1 to 4.0 times fewer requests at 64 clients, as it runs without CUDA graphs and builds the metadata of its Gated DeltaNet branches on the CPU.

\input{tables/resources}

\paragraph{Memory and GPU time.}
On Qwen3.5-4B in Table~\ref{tab:resources}, which runs the vLLM build of 14 October, vLLM and SGLang reserve most of the card for their key and value caches by default, and their peak follows a setting.
Flide holds 18.8~GiB at its peak, the model, 2.1~GiB of CUDA graphs and the work space of its largest rounds.
At 64 clients on DI every engine keeps a kernel running 71 to 91 percent of the time, yet Flide turns that time into 68 thousand useful tokens per second against 32 to 46 thousand for the others, since vLLM and SGLang spend part of it on states that they compute again for each question and multi-item scoring runs its kernels without CUDA graphs.
Busy time thus measures how full the GPU is, and useful tokens measure how much of that work a decision needed.
Held to a tenth of the card, vLLM and SGLang peak at 17.2 and 18.7~GiB, close to Flide, and Flide's lead holds at equal memory.

\begin{figure}[t]
\centering
\includegraphics[width=\linewidth]{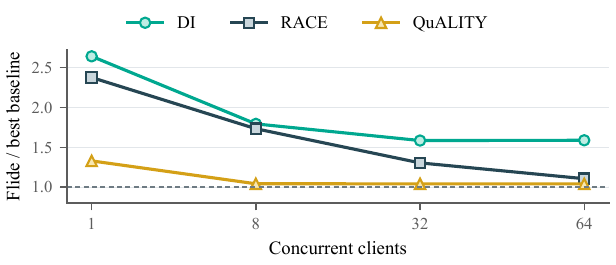}
\caption{The lead under load. Flide over the best baseline in throughput on Qwen3.5, mean over 0.8B, 4B and 9B.}
\label{fig:concurrency}
\end{figure}

\paragraph{The lead under load.}
As shown in Figure~\ref{fig:concurrency}, the lead of Flide changes with the number of clients.
With few clients the GPU waits for the engine, and removing launches and scheduling rounds pays off on every workload.
With many clients the single-question workloads become bound by the forward pass, which every engine computes in about the same time, and Flide still leads on QuALITY by about four percent.
On DI the lead persists at every level, because computing each state once saves GPU work as well as overhead.

  \begin{table*}[t]
  \begin{minipage}[b]{0.485\textwidth}\renewenvironment{table}[1][]{\centering}{}\input{tables/gqa}\end{minipage}\hfill
  \begin{minipage}[b]{0.485\textwidth}\renewenvironment{table}[1][]{\centering}{}\input{tables/repeat}\end{minipage}
  \end{table*}

\paragraph{Questions about images.}
Qwen3.5 reads images as well, and an agent that acts on a screen asks its questions about what it sees.
We take the 397 images of the balanced test-dev split of GQA \citep{hudson2019gqa} with their 4,525 yes-or-no questions, eleven per image on average, and send each image with all of its questions as one request.
This comparison runs the vLLM build of 14 October, whose decision endpoint takes no images for these models, and SGLang's endpoint takes none either, so both answer through their chat endpoints with one request per question, the image first and the label probabilities read from the first output token.
As Table~\ref{tab:gqa} shows, Flide answers an image 5.0 to 7.5 times faster than the better of the two alone and serves 3.1 to 5.4 times its requests at 64 clients, while its answers stay closest to the float32 reference on every model and all three engines stay within half a point of the reference's accuracy of 70.4, 77.7 and 79.3 percent.

\input{tables/startup}

\paragraph{Start-up.}
A Triton kernel is compiled on first use, and its tuning runs on first use as well, so an engine built on Triton can spend minutes before its first answer.
Flide ships its kernels compiled and tuned in a bundle independent of the Triton installation.
As Table~\ref{tab:startup} shows, a 0.8B server answers its first request after about 35 s with the bundle against 181 s without it, since the bundle brings every kernel compiled and tuned, and its answers are identical bit for bit.

\section{Recurring States}
\label{app:repeat}

Table~\ref{tab:repeat} sends every row once with every engine's cache on, Flide's saved states and the prefix caches of vLLM and SGLang, so a state that recurs can be served from the cache.
On the 4B model Flide answers a recurring QuALITY article in 18.4~ms against 39.1~ms for vLLM and 75.0~ms for SGLang, because it resumes from the saved recurrent state without recomputing a block.
Flide stays ahead in every cell, by 2.6 to 44 percent under load on the single-question workloads and by 1.6 to 2.1 times on DI.

\section{Correctness Tests}
\label{app:tests}

Every change to the engine passes a ladder of tests before any speed is measured.
The kernel tests compare the attention and fused pointwise kernels with a PyTorch \citep{paszke2019pytorch} reference over head sizes, grouped heads, sliding windows and every layout of a round, where every error stays at the rounding level of float32 and bfloat16 and the key blocks read without a mask return the same bits as with one.
The forward tests run OneJev1.5-0.8B \citep{onejev2026} in float32 on ten requests of DI \citep{decisionindex2026} against a float64 forward pass, where the transformers library \citep{wolf2020transformers} reaches a largest error of $6.0 \times 10^{-3}$ and every change of Flide must stay within twice that error.
Flide's eager forward and its CUDA graphs, requests replayed together in one flat graph, requests with images and videos and requests resumed from a saved state all stay within this bound, while a request replayed after a different one and a server with two front-end processes return the same results as before.

\section{Supported Models}
\label{app:models}

\input{tables/models}
Table~\ref{tab:models} lists the decision models that Flide serves, each against its release engine on the first 32 requests of RACE, MultiRC and QuALITY, of a 760-request sample of the Decision Index and of GQA with images.
Both mammoth and elephant read each block both ways in their full-attention layers, and since their state never reads a question, their questions still fork it under the rule of Section~\ref{sec:tree} without $t_y \le t_x$.
Jeffan, StartLux and Lychee read the whole prompt of each question both ways, so every question runs as a sequence of its own in the stream, and pplx-decider reads all questions of a request in one sequence.
Intern-Decision reads at most 16 questions a request, so its cells keep the requests within that limit among the first 32.
The Open-Jev row repeats RACE and QuALITY of Table~\ref{tab:vllmjev}, all 256 requests against the better setting of vllm-jev, and adds DI and MultiRC on their first 32 requests against the same setting.
The OneJev row compares OneJev1.5-0.8B with qev, the engine it ran on before Flide, on all 760 requests of the DI sample and the first 32 of MultiRC and QuALITY.
Models built on Qwen3.8, which shares the architecture of Qwen3.5, appear in the table with a Qwen3.5 backbone.

\end{document}

%% file: figures/teaser.tex
\begin{figure*}[t]
\centering
\begin{tikzpicture}[x=0.8mm, y=1mm, font=\small,
  st/.style={draw=stateD, fill=stateL, line width=0.5pt},
  ca/.style={draw=stateD, fill=stateL!35, line width=0.5pt},
  re/.style={draw=costD, fill=costL, line width=0.5pt},
  de/.style={draw=costD, fill=costD, line width=0.5pt},
  qa/.style={draw=qaD, fill=qaL, line width=0.5pt},
  qb/.style={draw=qbD, fill=qbL, line width=0.5pt},
  qc/.style={draw=qcD, fill=qcL, line width=0.5pt},
  title/.style={anchor=base, font=\small\bfseries},
  sub/.style={anchor=base west, font=\footnotesize}]
  \path (-1.5, 0);
  \node[title] at (26.43, 28.15) {Generation engine};
  \foreach \r/\q in {0/qa, 1/qb, 2/qc} {
    \pgfmathsetmacro{\y}{20 - 6.5*\r}
    \ifnum\r=0
      \draw[st] (0, \y) rectangle (41.25, \y+4.5);
    \else
      \draw[ca] (0, \y) rectangle (34.38, \y+4.5);
      \draw[re] (34.38, \y) rectangle (41.25, \y+4.5);
    \fi
    \foreach \b in {11.46, 22.92, 34.38} { \draw[stateD, line width=0.6pt] (\b, \y-0.7) -- (\b, \y+5.2); }
    \draw[\q] (41.25, \y) rectangle (47.25, \y+4.5);
    \pgfmathtruncatemacro{\k}{\r+1}
    \node[font=\footnotesize] at (44.25, \y+2.25) {$q_{\k}$};
    \node[anchor=west] at (47.75, \y+2.25) {$p_{\k}$};
  }
  \draw[st] (0, -0.5) rectangle (3, 2); \node[sub] at (3.5, 0.1) {computed};
  \draw[ca] (26, -0.5) rectangle (29, 2); \node[sub] at (29.5, 0.1) {from cache};
  \draw[re] (0, -5.55) rectangle (3, -3.05); \node[sub] at (3.5, -4.95) {recomputed};
  \draw[stateD, line width=0.6pt] (27.5, -6.2) -- (27.5, -2.4); \node[sub] at (29.5, -4.95) {saved block};
  \begin{scope}[shift={(72.66, 0)}]
    \node[title] at (23.76, 28.15) {Multi-item scoring};
    \draw[st] (0, 13.5) rectangle (28, 18);
    \node at (14, 15.75) {state $s$};
    \foreach \y/\q/\c/\k in {22/qa/qaD/1, 15.75/qb/qbD/2, 9.5/qc/qcD/3} {
      \draw[\c, line width=1pt] (28, 15.75) .. controls (31.5, 15.75) and (31, \y) .. (34.5, \y);
      \draw[de] (34.5, \y-2.25) rectangle (36.1, \y+2.25);
      \draw[\q] (36.1, \y-2.25) rectangle (42.1, \y+2.25);
      \node[font=\footnotesize] at (39.1, \y) {$q_{\k}$};
      \node[anchor=west, text=costD] at (42.6, \y) {$\tilde{p}_{\k}$};
    }
    \fill[stateD] (28, 15.75) circle (0.6mm);
    \draw[st] (0, -1) rectangle (16, 2.5); \node[font=\footnotesize] at (8, 0.75) {$s$};
    \foreach \x/\q/\k in {16/qa/1, 22.4/qb/2, 28.8/qc/3} {
      \draw[de] (\x, -1) rectangle (\x+1.4, 2.5);
      \draw[\q] (\x+1.4, -1) rectangle (\x+6.4, 2.5);
      \node[font=\scriptsize] at (\x+3.9, 0.75) {$q_{\k}$};
    }
    \node[sub] at (36.2, 0.1) {one sequence};
    \node[sub, anchor=base] at (24.27, -4.95) {\textcolor{costD}{\rule[-1.25mm]{1.3mm}{3.8mm}}\hspace{1.3mm}delimiter};
  \end{scope}
  \begin{scope}[shift={(140.1, 0)}]
    \node[title] at (23.08, 28.15) {Flide};
    \draw[st] (0, 13.5) rectangle (28, 18);
    \node at (14, 15.75) {state $s$};
    \foreach \y/\q/\c/\k in {22/qa/qaD/1, 15.75/qb/qbD/2, 9.5/qc/qcD/3} {
      \draw[\c, line width=1pt] (28, 15.75) .. controls (31.5, 15.75) and (31, \y) .. (34.5, \y);
      \draw[\q] (34.5, \y-2.25) rectangle (40.5, \y+2.25);
      \node[font=\footnotesize] at (37.5, \y) {$q_{\k}$};
      \node[anchor=west] at (41, \y) {$p_{\k}$};
    }
    \fill[stateD] (28, 15.75) circle (0.6mm);
    \draw[st] (0, -1) rectangle (16, 2.5); \node[font=\footnotesize] at (8, 0.75) {$s$};
    \foreach \x/\q/\k in {16/qa/1, 22/qb/2, 28/qc/3} {
      \draw[\q] (\x, -1) rectangle (\x+6, 2.5);
      \node[font=\scriptsize] at (\x+3, 0.75) {$q_{\k}$};
    }
    \node[sub] at (35.5, 0.1) {one stream};
    \foreach \x/\w/\c/\k in {0/16/stateD/0, 16/6/qaD/1, 22/6/qbD/2, 28/6/qcD/3} {
      \draw[\c, line width=0.6pt, fill=white] (\x+0.3, -6.2) rectangle (\x+\w-0.3, -2.4);
      \node[font=\footnotesize, text=\c] at (\x+\w/2, -4.3) {\k};
      \draw[\c, line width=0.5pt] (\x+\w/2, -2.4) -- (\x+\w/2, -1);
    }
    \node[sub] at (35.5, -4.95) {branch tags};
  \end{scope}
\end{tikzpicture}
\caption{One request with three questions, run three ways. Forks show the logical structure and strips the physical sequence, where red bars are delimiters and a tilde, as in $\tilde{p}_1$, marks an answer that may differ.}
\label{fig:teaser}
\end{figure*}

%% file: tables/engines.tex
\begin{table*}[t]
\centering
\tabsetup
\begin{tabular}{lccccc}
\toprule
\textbf{Engine} & \textbf{One pass} & \textbf{Same input} & \textbf{Hybrid fork} & \textbf{Graph replay} & \textbf{Bitwise} \\
\midrule
vLLM \citep{vllm2026systemone} & \xmark & \cmark & \xmark & \cmark & \xmark \\
SGLang \citep{sglang2026systemone} & \xmark & \cmark & \xmark & \cmark & \cmark \\
SGLang MIS \citep{sglang2026mis} & \cmark & \xmark & \cmark & \xmark & \cmark \\
vllm-jev \citep{modeio2026vllmjev} & \xmark & \cmark & \xmark & \cmark & \xmark \\
\textbf{Flide} & \cmark & \cmark & \cmark & \cmark & \cmark \\
\bottomrule
\end{tabular}
\caption{Decision serving in current engines. Each mark follows the public documentation and code of the engine, and bitwise marks an optional mode that keeps answers bit-identical under any load on hybrid models.}
\label{tab:engines}
\end{table*}

%% file: figures/method.tex
\begin{figure*}[!t]
\centering
\includegraphics[width=\textwidth]{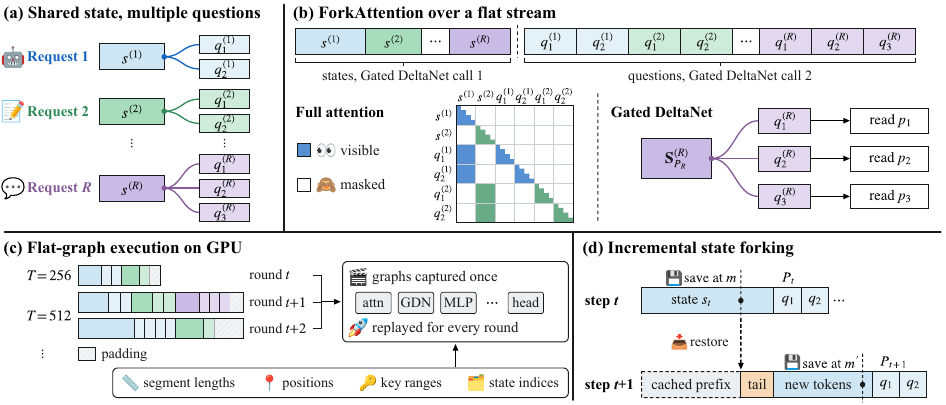}
\caption{Overview of Flide. (a) Requests, each a tree of a shared state and its questions. (b) ForkAttention over one flat stream. (c) Flat graphs over GPU-resident round metadata. (d) Incremental forks of a growing state.}
\label{fig:method}
\end{figure*}

%% file: tables/workloads.tex
\begin{table*}[t]
\centering
\tabsetup
\fontsize{9}{11}\selectfont\setlength{\tabcolsep}{3.4pt}
\begin{tabular}{ll rrrrrrrr}
\toprule
\textbf{Workload} & \textbf{Source} & Requests & Questions & Per request & State & Stream & Sequences & Copies & Per level \\
\midrule
RACE & RACE-M test & 256 & 256 & 1.0 & 250 & 325 & 325 & 4 & 1,024 \\
QuALITY & QuALITY val., cut & 256 & 256 & 1.0 & 3,355 & 3,443 & 3,443 & 4 & 1,024 \\
\hdashline\noalign{\vskip 1.5pt}
DI & Decision Index & 679 & 3,635 & 5.4 & 323 & 1,349 & 3,387 & 2 & 1,358 \\
RACE, all & RACE-M test & 256 & 1,019 & 4.0 & 255 & 546 & 1,283 & 4 & 1,024 \\
QuALITY, all & QuALITY val. & 115 & 2,086 & 18.1 & 5,835 & 7,454 & 107,085 & 2 & 230 \\
MultiRC & MultiRC val. & 83 & 3,948 & 47.6 & 327 & 2,663 & 17,808 & 4 & 332 \\
\bottomrule
\end{tabular}
\caption{Workloads. Mean tokens of the state, of the stream with the state once and every question once, and of one
sequence per question, in the prompts of Flide and vLLM.}
\label{tab:workloads}
\end{table*}

%% file: tables/main.tex
\begin{table*}[t]
\centering
\tabsetup
\fontsize{9}{11}\selectfont\setlength{\tabcolsep}{2pt}
\begin{tabular}{l@{\hspace{\tabcolsep}\hspace{\dimexpr(\textwidth-466.8pt)/7\relax}\hspace{\tabcolsep}}l@{\hspace{\tabcolsep}\hspace{\dimexpr(\textwidth-466.8pt)/7\relax}\hspace{\tabcolsep}}rrr@{\hspace{\tabcolsep}\hspace{\dimexpr(\textwidth-466.8pt)/7\relax}\hspace{\tabcolsep}}rrr@{\hspace{\tabcolsep}\hspace{\dimexpr(\textwidth-466.8pt)/7\relax}\hspace{\tabcolsep}}rrr@{\hspace{\tabcolsep}\hspace{\dimexpr(\textwidth-466.8pt)/7\relax}\hspace{\tabcolsep}}rrr@{\hspace{\tabcolsep}\hspace{\dimexpr(\textwidth-466.8pt)/7\relax}\hspace{\tabcolsep}}rrr@{\hspace{\tabcolsep}\hspace{\dimexpr(\textwidth-466.8pt)/7\relax}\hspace{\tabcolsep}}rrr}
\toprule
& & \multicolumn{6}{c}{\textbf{One question per state}} & \multicolumn{12}{c}{\textbf{All questions of a state}} \\
\cmidrule(lr){3-8}\cmidrule(lr){9-20}
& & \multicolumn{3}{c}{RACE} & \multicolumn{3}{c}{QuALITY} & \multicolumn{3}{c}{DI} & \multicolumn{3}{c}{RACE} & \multicolumn{3}{c}{QuALITY} & \multicolumn{3}{c}{MultiRC} \\
\cmidrule(lr){3-5}\cmidrule(lr){6-8}\cmidrule(lr){9-11}\cmidrule(lr){12-14}\cmidrule(lr){15-17}\cmidrule(lr){18-20}
\textbf{Model} & \textbf{Engine} & Lat.$\downarrow$ & C8$\uparrow$ & C64$\uparrow$ & Lat.$\downarrow$ & C8$\uparrow$ & C64$\uparrow$ & Lat.$\downarrow$ & C8$\uparrow$ & C64$\uparrow$ & Lat.$\downarrow$ & C8$\uparrow$ & C64$\uparrow$ & Lat.$\downarrow$ & C8$\uparrow$ & C64$\uparrow$ & Lat.$\downarrow$ & C8$\uparrow$ & C64$\uparrow$ \\
\midrule
\multicolumn{20}{l}{\textbf{Qwen3.5}} \\
\multirow{3}{*}{0.8B} & vLLM & 19.9 & 154 & 527 & 35.1 & 84.1 & 89.4 & 23.2 & 69.1 & 107 & 47.3 & 88.2 & 208 & 362 & 4.5 & 4.6 & 154 & 14.1 & 14.4 \\
 & SGLang & 27.0 & 153 & 418 & 35.3 & 76.5 & 78.8 & 28.6 & 61.5 & 98.2 & 31.0 & 95.7 & 174 & 359 & 4.6 & 4.4 & 130 & 16.6 & 15.1 \\
 & Flide & \cellcolor{tabaccent!45}\textbf{6.2} & \cellcolor{tabaccent!45}\textbf{539} & \cellcolor{tabaccent!18}\textbf{672} & \cellcolor{tabaccent!37}\textbf{17.9} & \cellcolor{tabaccent!13}\textbf{93.6} & \cellcolor{tabaccent!11}\textbf{96.5} & \cellcolor{tabaccent!45}\textbf{7.1} & \cellcolor{tabaccent!45}\textbf{196} & \cellcolor{tabaccent!42}\textbf{236} & \cellcolor{tabaccent!45}\textbf{7.2} & \cellcolor{tabaccent!45}\textbf{410} & \cellcolor{tabaccent!45}\textbf{540} & \cellcolor{tabaccent!45}\textbf{37.7} & \cellcolor{tabaccent!45}\textbf{39.1} & \cellcolor{tabaccent!45}\textbf{39.1} & \cellcolor{tabaccent!45}\textbf{16.8} & \cellcolor{tabaccent!45}\textbf{115} & \cellcolor{tabaccent!45}\textbf{124} \\
\hdashline\noalign{\vskip 1.5pt}
\multirow{3}{*}{4B} & vLLM & 24.6 & 130 & 196 & 68.6 & 19.3 & 19.3 & 27.7 & 31.6 & 36.0 & 57.6 & 49.2 & 67.0 & 476 & 4.3 & 4.2 & 281 & 7.1 & 4.9 \\
 & SGLang & 31.6 & 96.5 & 179 & 75.3 & 18.5 & 18.5 & 34.8 & 24.0 & 27.9 & 37.4 & 45.4 & 49.8 & 727 & 2.4 & 2.1 & 272 & 5.2 & 5.3 \\
 & Flide & \cellcolor{tabaccent!43}\textbf{10.9} & \cellcolor{tabaccent!21}\textbf{175} & \cellcolor{tabaccent!10}\textbf{204} & \cellcolor{tabaccent!12}\textbf{62.3} & \cellcolor{tabaccent!8}\textbf{19.4} & \cellcolor{tabaccent!9}\textbf{19.9} & \cellcolor{tabaccent!45}\textbf{11.9} & \cellcolor{tabaccent!26}\textbf{48.3} & \cellcolor{tabaccent!23}\textbf{50.4} & \cellcolor{tabaccent!45}\textbf{14.1} & \cellcolor{tabaccent!45}\textbf{116} & \cellcolor{tabaccent!35}\textbf{124} & \cellcolor{tabaccent!45}\textbf{141} & \cellcolor{tabaccent!38}\textbf{8.6} & \cellcolor{tabaccent!39}\textbf{8.6} & \cellcolor{tabaccent!45}\textbf{50.3} & \cellcolor{tabaccent!45}\textbf{25.5} & \cellcolor{tabaccent!45}\textbf{25.9} \\
\hdashline\noalign{\vskip 1.5pt}
\multirow{3}{*}{9B} & vLLM & 24.7 & 96.0 & 117 & 106 & 11.3 & 11.3 & 26.0 & 21.4 & 22.7 & 60.1 & 38.4 & 46.4 & 562 & 3.0 & 3.5 & 308 & 5.8 & 3.1 \\
 & SGLang & 32.6 & 80.4 & 108 & 117 & 10.5 & 10.9 & 33.2 & 14.8 & 16.2 & 50.9 & 26.4 & 28.9 & 1.1k & 1.5 & 1.4 & 416 & 3.1 & 3.1 \\
 & Flide & \cellcolor{tabaccent!34}\textbf{13.7} & \cellcolor{tabaccent!12}\textbf{106} & \cellcolor{tabaccent!9}\textbf{119} & \cellcolor{tabaccent!10}\textbf{102} & \cellcolor{tabaccent!8}\textbf{11.5} & \cellcolor{tabaccent!9}\textbf{11.5} & \cellcolor{tabaccent!34}\textbf{14.3} & \cellcolor{tabaccent!20}\textbf{28.4} & \cellcolor{tabaccent!19}\textbf{29.2} & \cellcolor{tabaccent!45}\textbf{20.4} & \cellcolor{tabaccent!32}\textbf{67.1} & \cellcolor{tabaccent!26}\textbf{71.0} & \cellcolor{tabaccent!45}\textbf{232} & \cellcolor{tabaccent!32}\textbf{5.1} & \cellcolor{tabaccent!25}\textbf{5.1} & \cellcolor{tabaccent!45}\textbf{81.1} & \cellcolor{tabaccent!45}\textbf{14.7} & \cellcolor{tabaccent!45}\textbf{14.9} \\
\hdashline\noalign{\vskip 1.5pt}
\multirow{3}{*}{27B} & vLLM & 48.2 & 31.4 & 34.5 & \textbf{337} & \textbf{3.3} & \textbf{3.3} & 48.7 & 6.6 & 5.7 & 140 & 12.6 & 13.0 & 1.2k & 1.3 & 0.8 & 656 & 1.9 & 0.6 \\
 & SGLang & 55.9 & 27.9 & 32.6 & 363 & 3.2 & 3.2 & 57.1 & 4.6 & 4.8 & 134 & 8.0 & 8.4 & 3.3k & 0.4 & 0.4 & 1.1k & 1.0 & 1.0 \\
 & Flide & \cellcolor{tabaccent!15}\textbf{41.2} & \cellcolor{tabaccent!9}\textbf{32.3} & \cellcolor{tabaccent!8}\textbf{34.8} & 340 & \textbf{3.3} & \textbf{3.3} & \cellcolor{tabaccent!15}\textbf{41.1} & \cellcolor{tabaccent!18}\textbf{8.3} & \cellcolor{tabaccent!26}\textbf{8.5} & \cellcolor{tabaccent!39}\textbf{66.1} & \cellcolor{tabaccent!28}\textbf{19.9} & \cellcolor{tabaccent!29}\textbf{21.0} & \cellcolor{tabaccent!25}\textbf{781} & \cellcolor{tabaccent!14}\textbf{1.5} & \cellcolor{tabaccent!38}\textbf{1.5} & \cellcolor{tabaccent!45}\textbf{270} & \cellcolor{tabaccent!44}\textbf{4.3} & \cellcolor{tabaccent!45}\textbf{4.3} \\
\midrule
\multirow{3}{*}{Average} & vLLM & 29.4 & 103 & 219 & 137 & 29.5 & 30.8 & 31.4 & 32.2 & 42.7 & 76.2 & 47.1 & 83.7 & 639 & 3.3 & 3.2 & 350 & 7.2 & 5.8 \\
 & SGLang & 36.8 & 89.4 & 184 & 148 & 27.2 & 27.9 & 38.4 & 26.2 & 36.8 & 63.3 & 43.9 & 65.2 & 1.4k & 2.2 & 2.1 & 491 & 6.5 & 6.1 \\
 & Flide & \cellcolor{tabaccent!29}\textbf{18.0} & \cellcolor{tabaccent!40}\textbf{213} & \cellcolor{tabaccent!15}\textbf{257} & \cellcolor{tabaccent!10}\textbf{130} & \cellcolor{tabaccent!11}\textbf{32.0} & \cellcolor{tabaccent!11}\textbf{32.8} & \cellcolor{tabaccent!31}\textbf{18.6} & \cellcolor{tabaccent!42}\textbf{70.3} & \cellcolor{tabaccent!36}\textbf{81.1} & \cellcolor{tabaccent!45}\textbf{26.9} & \cellcolor{tabaccent!45}\textbf{153} & \cellcolor{tabaccent!43}\textbf{189} & \cellcolor{tabaccent!41}\textbf{298} & \cellcolor{tabaccent!45}\textbf{13.6} & \cellcolor{tabaccent!45}\textbf{13.6} & \cellcolor{tabaccent!45}\textbf{105} & \cellcolor{tabaccent!45}\textbf{39.9} & \cellcolor{tabaccent!45}\textbf{42.4} \\
\midrule
\multicolumn{20}{l}{\textbf{Gemma 4}} \\
\multirow{3}{*}{E2B} & vLLM & 12.9 & 186 & 353 & 41.9 & \textbf{32.5} & \textbf{32.6} & 13.4 & 63.7 & 77.7 & 24.3 & 107 & 201 & 255 & 7.5 & 7.1 & 70.4 & 26.4 & 34.5 \\
 & SGLang & 32.7 & 100 & 303 & 66.6 & 20.5 & 17.6 & 33.7 & 35.2 & 44.8 & 37.0 & 58.8 & 82.2 & 549 & 2.8 & 2.5 & 185 & 9.9 & 9.4 \\
 & Flide & \cellcolor{tabaccent!28}\textbf{8.2} & \cellcolor{tabaccent!31}\textbf{314} & \cellcolor{tabaccent!9}\textbf{360} & \cellcolor{tabaccent!14}\textbf{36.5} & 32.3 & 32.5 & \cellcolor{tabaccent!26}\textbf{8.8} & \cellcolor{tabaccent!20}\textbf{83.8} & \cellcolor{tabaccent!13}\textbf{87.6} & \cellcolor{tabaccent!44}\textbf{10.7} & \cellcolor{tabaccent!35}\textbf{200} & \cellcolor{tabaccent!13}\textbf{224} & \cellcolor{tabaccent!45}\textbf{86.6} & \cellcolor{tabaccent!34}\textbf{13.8} & \cellcolor{tabaccent!37}\textbf{13.7} & \cellcolor{tabaccent!45}\textbf{28.3} & \cellcolor{tabaccent!34}\textbf{48.6} & \cellcolor{tabaccent!24}\textbf{49.6} \\
\hdashline\noalign{\vskip 1.5pt}
\multirow{3}{*}{E4B} & vLLM & 14.5 & 154 & 192 & 65.7 & \textbf{18.0} & \textbf{18.0} & 14.8 & 39.7 & 42.6 & 25.8 & 95.4 & 109 & 364 & 4.3 & 4.2 & 87.2 & 23.3 & 23.8 \\
 & SGLang & 36.6 & 105 & 173 & 95.3 & 13.2 & 13.2 & 38.3 & 23.1 & 26.7 & 41.5 & 37.7 & 45.4 & 708 & 1.9 & 1.8 & 265 & 5.6 & 5.5 \\
 & Flide & \cellcolor{tabaccent!19}\textbf{11.3} & \cellcolor{tabaccent!13}\textbf{172} & \cellcolor{tabaccent!8}\textbf{193} & \cellcolor{tabaccent!9}\textbf{63.8} & 17.8 & 17.9 & \cellcolor{tabaccent!19}\textbf{11.6} & \cellcolor{tabaccent!14}\textbf{45.5} & \cellcolor{tabaccent!12}\textbf{47.1} & \cellcolor{tabaccent!31}\textbf{15.3} & \cellcolor{tabaccent!13}\textbf{106} & \cellcolor{tabaccent!11}\textbf{117} & \cellcolor{tabaccent!45}\textbf{148} & \cellcolor{tabaccent!33}\textbf{7.8} & \cellcolor{tabaccent!35}\textbf{7.8} & \cellcolor{tabaccent!32}\textbf{50.2} & \cellcolor{tabaccent!12}\textbf{25.3} & \cellcolor{tabaccent!11}\textbf{25.7} \\
\hdashline\noalign{\vskip 1.5pt}
\multirow{3}{*}{12B} & vLLM & 22.4 & 68.2 & \textbf{75.8} & \textbf{157} & \textbf{7.0} & \textbf{7.0} & 23.4 & 15.8 & 16.1 & 40.8 & 36.8 & 38.8 & 763 & 1.6 & 1.6 & 191 & 7.2 & 7.1 \\
 & SGLang & 36.5 & 60.1 & 69.9 & 218 & 5.2 & 5.1 & 38.9 & 9.2 & 9.7 & 65.8 & 16.6 & 18.1 & 1.5k & 0.8 & 0.7 & 557 & 2.2 & 2.2 \\
 & Flide & \cellcolor{tabaccent!13}\textbf{19.9} & \cellcolor{tabaccent!9}\textbf{69.8} & 75.4 & 158 & \textbf{7.0} & \textbf{7.0} & \cellcolor{tabaccent!16}\textbf{19.4} & \cellcolor{tabaccent!13}\textbf{17.9} & \cellcolor{tabaccent!14}\textbf{18.3} & \cellcolor{tabaccent!22}\textbf{29.7} & \cellcolor{tabaccent!15}\textbf{42.9} & \cellcolor{tabaccent!14}\textbf{45.1} & \cellcolor{tabaccent!39}\textbf{375} & \cellcolor{tabaccent!34}\textbf{3.0} & \cellcolor{tabaccent!35}\textbf{3.0} & \cellcolor{tabaccent!26}\textbf{125} & \cellcolor{tabaccent!19}\textbf{9.4} & \cellcolor{tabaccent!21}\textbf{9.5} \\
\hdashline\noalign{\vskip 1.5pt}
\multirow{3}{*}{31B} & vLLM & 46.9 & \textbf{27.7} & \textbf{30.1} & \textbf{400} & \textbf{2.7} & 2.6 & 45.2 & 6.2 & 6.2 & 83.1 & 15.8 & 16.0 & 1.8k & 0.6 & 0.6 & 347 & 3.5 & 3.5 \\
 & SGLang & 52.2 & 24.7 & 26.8 & 543 & 1.9 & 1.9 & 52.7 & 3.6 & 3.7 & 152 & 6.6 & 6.8 & 3.7k & 0.3 & 0.3 & 1.4k & 0.8 & 0.8 \\
 & Flide & \cellcolor{tabaccent!10}\textbf{44.4} & 27.4 & 29.2 & 405 & \textbf{2.7} & \cellcolor{tabaccent!8}\textbf{2.7} & \cellcolor{tabaccent!11}\textbf{41.9} & \cellcolor{tabaccent!13}\textbf{6.9} & \cellcolor{tabaccent!13}\textbf{7.0} & \cellcolor{tabaccent!17}\textbf{68.2} & \cellcolor{tabaccent!10}\textbf{16.6} & \cellcolor{tabaccent!11}\textbf{17.4} & \cellcolor{tabaccent!35}\textbf{982} & \cellcolor{tabaccent!33}\textbf{1.1} & \cellcolor{tabaccent!34}\textbf{1.1} & \cellcolor{tabaccent!12}\textbf{319} & \cellcolor{tabaccent!9}\textbf{3.6} & \cellcolor{tabaccent!10}\textbf{3.6} \\
\midrule
\multirow{3}{*}{Average} & vLLM & 24.2 & 109 & 163 & \textbf{166} & \textbf{15.1} & \textbf{15.1} & 24.2 & 31.3 & 35.6 & 43.5 & 63.7 & 91.2 & 804 & 3.5 & 3.4 & 174 & 15.1 & 17.2 \\
 & SGLang & 39.5 & 72.4 & 143 & 231 & 10.2 & 9.4 & 40.9 & 17.8 & 21.2 & 74.2 & 29.9 & 38.1 & 1.6k & 1.5 & 1.3 & 594 & 4.6 & 4.5 \\
 & Flide & \cellcolor{tabaccent!14}\textbf{20.9} & \cellcolor{tabaccent!21}\textbf{146} & \cellcolor{tabaccent!8}\textbf{164} & \cellcolor{tabaccent!8}\textbf{166} & 14.9 & 15.0 & \cellcolor{tabaccent!15}\textbf{20.4} & \cellcolor{tabaccent!17}\textbf{38.5} & \cellcolor{tabaccent!13}\textbf{40.0} & \cellcolor{tabaccent!23}\textbf{31.0} & \cellcolor{tabaccent!24}\textbf{91.4} & \cellcolor{tabaccent!12}\textbf{101} & \cellcolor{tabaccent!38}\textbf{398} & \cellcolor{tabaccent!34}\textbf{6.4} & \cellcolor{tabaccent!36}\textbf{6.4} & \cellcolor{tabaccent!20}\textbf{131} & \cellcolor{tabaccent!24}\textbf{21.7} & \cellcolor{tabaccent!19}\textbf{22.1} \\
\bottomrule
\end{tabular}
\caption{Flide leads most where a state carries many questions, and shading gives its gain over the best baseline.}
\label{tab:main}
\end{table*}

%% file: tables/fidelity.tex
\begin{table*}[t]
\centering
\tabsetup
\fontsize{8.7}{10.6}\selectfont\setlength{\tabcolsep}{1.1pt}
\begin{tabular}{l@{\hspace{\tabcolsep}\hspace{\dimexpr(\textwidth-473.4pt)/7\relax}\hspace{\tabcolsep}}l@{\hspace{\tabcolsep}\hspace{\dimexpr(\textwidth-473.4pt)/7\relax}\hspace{\tabcolsep}}rrr@{\hspace{\tabcolsep}\hspace{\dimexpr(\textwidth-473.4pt)/7\relax}\hspace{\tabcolsep}}rrr@{\hspace{\tabcolsep}\hspace{\dimexpr(\textwidth-473.4pt)/7\relax}\hspace{\tabcolsep}}rr@{\hspace{\tabcolsep}\hspace{\dimexpr(\textwidth-473.4pt)/7\relax}\hspace{\tabcolsep}}rrr@{\hspace{\tabcolsep}\hspace{\dimexpr(\textwidth-473.4pt)/7\relax}\hspace{\tabcolsep}}rrr@{\hspace{\tabcolsep}\hspace{\dimexpr(\textwidth-473.4pt)/7\relax}\hspace{\tabcolsep}}rrr}
\toprule
& & \multicolumn{6}{c}{\textbf{One question per state}} & \multicolumn{11}{c}{\textbf{All questions of a state}} \\
\cmidrule(lr){3-8}\cmidrule(lr){9-19}
& & \multicolumn{3}{c}{RACE} & \multicolumn{3}{c}{QuALITY} & \multicolumn{2}{c}{DI} & \multicolumn{3}{c}{RACE} & \multicolumn{3}{c}{QuALITY} & \multicolumn{3}{c}{MultiRC} \\
\cmidrule(lr){3-5}\cmidrule(lr){6-8}\cmidrule(lr){9-10}\cmidrule(lr){11-13}\cmidrule(lr){14-16}\cmidrule(lr){17-19}
\textbf{Qwen3.5} & \textbf{Engine} & Agr.$\uparrow$ & Acc.$\uparrow$ & Stab.$\uparrow$ & Agr.$\uparrow$ & Acc.$\uparrow$ & Stab.$\uparrow$ & Agr.$\uparrow$ & Stab.$\uparrow$ & Agr.$\uparrow$ & Acc.$\uparrow$ & Stab.$\uparrow$ & Agr.$\uparrow$ & Acc.$\uparrow$ & Stab.$\uparrow$ & Agr.$\uparrow$ & Acc.$\uparrow$ & Stab.$\uparrow$ \\
\midrule
\multirow{3}{*}{0.8B} & vLLM & 98.0 & 66.4 & 97.8 & 98.0 & \textbf{39.5} & 96.8 & 97.8 & 99.0 & 97.6 & \textbf{71.7} & 97.9 & 98.0 & 46.1 & 99.1 & 99.7 & 44.4 & \textbf{100.0} \\
 & SGLang & 98.0 & \textbf{67.2} & 98.3 & 98.0 & \textbf{39.5} & 99.2 & 96.9 & 99.7 & 97.1 & 71.6 & 99.3 & 97.7 & \textbf{46.2} & 99.5 & 98.9 & \textbf{47.1} & 99.9 \\
 & Flide & \textbf{99.6} & \textbf{67.2} & \textbf{100.0} & \textbf{98.4} & 38.3 & \textbf{100.0} & \textbf{98.8} & \textbf{99.9} & \textbf{99.0} & \textbf{71.7} & \textbf{100.0} & \textbf{98.1} & 45.6 & \textbf{100.0} & \textbf{99.8} & 44.5 & \textbf{100.0} \\
\hdashline\noalign{\vskip 1.5pt}
\multirow{3}{*}{4B} & vLLM & \textbf{100.0} & 87.9 & 99.6 & 99.2 & 63.3 & 99.5 & 99.1 & 99.7 & \textbf{99.7} & 89.1 & 100.0 & 99.2 & 77.0 & 99.7 & 99.2 & 79.9 & \textbf{100.0} \\
 & SGLang & 99.6 & \textbf{88.3} & 99.6 & \textbf{99.6} & 63.3 & 99.7 & 99.3 & \textbf{99.8} & 99.6 & \textbf{89.2} & 100.0 & 99.3 & \textbf{77.5} & 99.5 & 98.3 & 74.9 & 99.9 \\
 & Flide & 99.6 & 87.5 & \textbf{100.0} & 98.8 & 63.3 & \textbf{100.0} & \textbf{99.4} & \textbf{99.8} & 99.6 & \textbf{89.2} & 100.0 & \textbf{99.6} & 77.1 & \textbf{100.0} & \textbf{99.5} & \textbf{80.2} & \textbf{100.0} \\
\hdashline\noalign{\vskip 1.5pt}
\multirow{3}{*}{9B} & vLLM & \textbf{100.0} & \textbf{92.6} & 99.8 & 98.4 & 70.3 & 99.7 & 99.4 & 99.6 & 99.8 & 92.8 & 99.7 & 99.3 & \textbf{83.0} & 99.8 & 98.5 & 78.9 & 99.9 \\
 & SGLang & 99.2 & 92.2 & 99.9 & 98.8 & 70.3 & 99.2 & 99.4 & 99.7 & 99.8 & 92.8 & 99.8 & 99.5 & \textbf{83.0} & 99.6 & 99.0 & 78.2 & 99.9 \\
 & Flide & \textbf{100.0} & \textbf{92.6} & \textbf{100.0} & \textbf{99.6} & 70.3 & \textbf{100.0} & \textbf{99.6} & \textbf{100.0} & 99.8 & \textbf{93.2} & \textbf{100.0} & \textbf{99.8} & 82.8 & \textbf{100.0} & \textbf{99.5} & \textbf{79.8} & \textbf{100.0} \\
\hdashline\noalign{\vskip 1.5pt}
\multirow{3}{*}{27B} & vLLM & 100.0 & 94.5 & \textbf{100.0} & 98.8 & 78.9 & 99.6 & 99.7 & 99.8 & 99.9 & \textbf{95.1} & 100.0 & 99.6 & 90.8 & 99.7 & 99.7 & 89.8 & \textbf{100.0} \\
 & SGLang & 100.0 & 94.5 & 99.9 & 98.8 & 78.9 & 99.5 & 99.4 & 99.8 & 99.8 & 95.0 & 100.0 & \textbf{99.8} & 90.8 & 99.7 & 99.7 & \textbf{90.5} & 99.9 \\
 & Flide & 100.0 & 94.5 & \textbf{100.0} & \textbf{99.6} & 78.9 & \textbf{100.0} & \textbf{99.9} & \textbf{100.0} & \textbf{100.0} & 95.0 & 100.0 & 99.7 & 90.8 & \textbf{100.0} & \textbf{99.9} & 89.8 & \textbf{100.0} \\
\midrule
\multirow{3}{*}{Average} & vLLM & 99.5 & 85.4 & 99.3 & 98.6 & \textbf{63.0} & 98.9 & 99.0 & 99.5 & 99.3 & 87.2 & 99.4 & 99.0 & 74.2 & 99.6 & 99.3 & 73.3 & \textbf{100.0} \\
 & SGLang & 99.2 & \textbf{85.5} & 99.4 & 98.8 & \textbf{63.0} & 99.4 & 98.7 & 99.7 & 99.1 & 87.2 & 99.8 & 99.1 & \textbf{74.4} & 99.6 & 99.0 & 72.7 & 99.9 \\
 & Flide & \textbf{99.8} & \textbf{85.5} & \textbf{100.0} & \textbf{99.1} & 62.7 & \textbf{100.0} & \textbf{99.4} & \textbf{99.9} & \textbf{99.6} & \textbf{87.3} & \textbf{100.0} & \textbf{99.3} & 74.1 & \textbf{100.0} & \textbf{99.7} & \textbf{73.6} & \textbf{100.0} \\
\bottomrule
\end{tabular}
\caption{On Qwen3.5, Flide's answers stay closest to a float32 reference, with all values in percent.}
\label{tab:fidelity}
\end{table*}

%% file: tables/ablation.tex
\begin{table}[t]
\centering
\tabsetup
\fontsize{9}{11}\selectfont\setlength{\tabcolsep}{1.7pt}
\begin{tabular}{l rr rr r}
\toprule
& \multicolumn{2}{c}{\textbf{DI}} & \multicolumn{2}{c}{\textbf{QuALITY, all}} & \textbf{RACE} \\
\cmidrule(lr){2-3}\cmidrule(lr){4-5}\cmidrule(lr){6-6}
\textbf{Variant} & Lat.$\downarrow$ & Req/s$\uparrow$ & Lat.$\downarrow$ & Req/s$\uparrow$ & Req/s$\uparrow$ \\
\midrule
Flide & 8.8 & \textbf{48.9} & \textbf{135.6} & \textbf{8.6} & 210.3 \\
without forks & 8.7 & 20.7 & 1868.5 & 0.6 & \textbf{211.1} \\
without flat graphs & \textbf{8.5} & 43.5 & 135.7 & 8.5 & 134.9 \\
without CUDA graphs & 76.1 & \textbf{48.9} & 137.2 & 8.5 & 191.6 \\
without selective readout & 8.9 & 47.8 & 140.1 & 8.3 & 205.2 \\
\bottomrule
\end{tabular}
\caption{Ablation on Qwen3.5-4B, timed inside the engine without its HTTP server. Each row removes one component of Flide, and Req/s counts the requests per second of a round.}
\label{tab:ablation}
\end{table}

%% file: tables/vllmjev.tex
\begin{table*}[t]
\centering
\tabsetup
\fontsize{9}{11}\selectfont\setlength{\tabcolsep}{5pt}
\begin{tabular}{l@{\hspace{\tabcolsep}\hspace{\dimexpr(\textwidth-439.9pt)/3\relax}\hspace{\tabcolsep}}l@{\hspace{\tabcolsep}\hspace{\dimexpr(\textwidth-439.9pt)/3\relax}\hspace{\tabcolsep}}rrrrr@{\hspace{\tabcolsep}\hspace{\dimexpr(\textwidth-439.9pt)/3\relax}\hspace{\tabcolsep}}rrrrr}
\toprule
& & \multicolumn{5}{c}{\textbf{RACE}} & \multicolumn{5}{c}{\textbf{QuALITY}} \\
\cmidrule(lr){3-7}\cmidrule(lr){8-12}
\textbf{Model} & \textbf{Engine} & Lat.$\downarrow$ & C8$\uparrow$ & C32$\uparrow$ & C64$\uparrow$ & Acc.$\uparrow$ & Lat.$\downarrow$ & C8$\uparrow$ & C32$\uparrow$ & C64$\uparrow$ & Acc.$\uparrow$ \\
\midrule
\multirow{3}{*}{Open-Jev-2B} & vllm-jev & 41.8 & 96.5 & 134.8 & 133.9 & \textbf{75.4} & 61.6 & 33.4 & 33.2 & 32.5 & 50.8 \\
 & + RadixFlow & 34.3 & 95.1 & 115.8 & 114.8 & 74.6 & 72.8 & 29.3 & 29.2 & 29.0 & 50.8 \\
 & Flide & \cellcolor{tabaccent!45}\textbf{8.9} & \cellcolor{tabaccent!45}\textbf{344.5} & \cellcolor{tabaccent!45}\textbf{401.4} & \cellcolor{tabaccent!45}\textbf{412.1} & 74.2 & \cellcolor{tabaccent!30}\textbf{37.3} & \cellcolor{tabaccent!23}\textbf{47.1} & \cellcolor{tabaccent!24}\textbf{47.9} & \cellcolor{tabaccent!25}\textbf{47.9} & 50.8 \\
\hdashline\noalign{\vskip 1.5pt}
\multirow{3}{*}{Open-Jev-9B} & vllm-jev & 57.8 & 31.8 & 32.9 & 32.9 & 84.4 & 141.9 & 8.5 & 8.5 & 8.5 & 68.4 \\
 & + RadixFlow & 47.5 & 30.4 & 31.7 & 31.7 & 84.4 & 153.9 & 8.2 & 8.1 & 8.2 & 68.4 \\
 & Flide & \cellcolor{tabaccent!45}\textbf{17.1} & \cellcolor{tabaccent!45}\textbf{93.5} & \cellcolor{tabaccent!45}\textbf{99.1} & \cellcolor{tabaccent!45}\textbf{100.7} & 84.4 & \cellcolor{tabaccent!20}\textbf{108.2} & \cellcolor{tabaccent!21}\textbf{11.3} & \cellcolor{tabaccent!21}\textbf{11.4} & \cellcolor{tabaccent!21}\textbf{11.4} & 68.4 \\
\midrule
\multirow{3}{*}{Average} & vllm-jev & 49.8 & 64.1 & 83.8 & 83.4 & \textbf{79.9} & 101.8 & 21.0 & 20.9 & 20.5 & 59.6 \\
 & + RadixFlow & 40.9 & 62.8 & 73.8 & 73.2 & 79.5 & 113.3 & 18.7 & 18.6 & 18.6 & 59.6 \\
 & Flide & \cellcolor{tabaccent!45}\textbf{13.0} & \cellcolor{tabaccent!45}\textbf{219.0} & \cellcolor{tabaccent!45}\textbf{250.3} & \cellcolor{tabaccent!45}\textbf{256.4} & 79.3 & \cellcolor{tabaccent!23}\textbf{72.8} & \cellcolor{tabaccent!22}\textbf{29.2} & \cellcolor{tabaccent!23}\textbf{29.7} & \cellcolor{tabaccent!24}\textbf{29.7} & 59.6 \\
\bottomrule
\end{tabular}
\caption{Flide against the engine written for these models. Shading gives Flide's gain over the better vllm-jev setting.}
\label{tab:vllmjev}
\end{table*}

%% file: tables/traj.tex
\begin{table*}[t]
\centering
\tabsetup
\fontsize{9}{11}\selectfont\setlength{\tabcolsep}{2.5pt}
\begin{tabular}{l@{\hspace{10pt}}rrrrr@{\hspace{10pt}}rrrrr@{\hspace{10pt}}rrrrr}
\toprule
& \multicolumn{5}{c}{\textbf{Qwen3.5-0.8B}} & \multicolumn{5}{c}{\textbf{Qwen3.5-4B}} & \multicolumn{5}{c}{\textbf{Qwen3.5-9B}} \\
\cmidrule(lr){2-6}\cmidrule(lr){7-11}\cmidrule(lr){12-16}
& \multicolumn{3}{c}{One agent, ms} & \multicolumn{2}{c}{Steps/s} & \multicolumn{3}{c}{One agent, ms} & \multicolumn{2}{c}{Steps/s} & \multicolumn{3}{c}{One agent, ms} & \multicolumn{2}{c}{Steps/s} \\
\cmidrule(lr){2-4}\cmidrule(lr){5-6}\cmidrule(lr){7-9}\cmidrule(lr){10-11}\cmidrule(lr){12-14}\cmidrule(lr){15-16}
\textbf{Engine} & ${<}$4k$\downarrow$ & ${>}$12k$\downarrow$ & Growth$\downarrow$ & 1$\uparrow$ & 32$\uparrow$ & ${<}$4k$\downarrow$ & ${>}$12k$\downarrow$ & Growth$\downarrow$ & 1$\uparrow$ & 32$\uparrow$ & ${<}$4k$\downarrow$ & ${>}$12k$\downarrow$ & Growth$\downarrow$ & 1$\uparrow$ & 32$\uparrow$ \\
\midrule
vLLM & 173.8 & 333.7 & 13.5 & 4.6 & 7.8 & 218.7 & 384.6 & 13.8 & 3.8 & 6.0 & 218.5 & 382.7 & 13.6 & 3.8 & 5.9 \\
SGLang & 64.2 & 197.5 & 11.1 & 9.7 & 14.0 & 116.2 & 279.8 & 13.2 & 6.0 & 12.4 & 167.6 & 355.5 & 14.5 & 4.5 & 8.2 \\
MIS & 52.4 & 88.3 & 2.9 & 16.4 & 28.9 & 81.8 & 294.9 & 17.2 & 7.5 & 8.9 & 102.2 & 453.8 & 28.5 & 5.2 & 5.7 \\
Flide & \textbf{11.7} & \textbf{18.7} & \textbf{0.6} & \textbf{72.4} & \textbf{208.2} & \textbf{22.4} & \textbf{36.1} & \textbf{0.9} & \textbf{38.5} & \textbf{57.9} & \textbf{32.7} & \textbf{48.2} & \textbf{1.0} & \textbf{27.4} & \textbf{36.1} \\
\bottomrule
\end{tabular}
\caption{Agents whose state grows by one step at a time. Latency of a step in ms with a history under 4k and over 12k tokens, its growth per thousand tokens of history, and steps per second at 1 and 32 agents.}
\label{tab:traj}
\end{table*}

%% file: tables/launch.tex
\begin{table*}[t]
\centering
\tabsetup
\begin{tabular}{l l >{\raggedright\arraybackslash}m{0.53\textwidth}}
\toprule
\textbf{Engine} & \textbf{Version} & \textbf{Endpoint and settings} \\
\midrule
vLLM & 0.32.0rc1 of 14 October, 0.32.1rc1 & decision endpoint, \texttt{-{}-enable-prefix-caching -{}-logprobs-mode raw\_logprobs}, the later build also \texttt{-{}-prefix-match-unit 32}; Gemma 4 also through the completion endpoint of 0.32.0rc1 \\
SGLang & 0.5.21 & decision endpoint, default settings \\
SGLang MIS & 0.5.21 & \texttt{/v1/score}, \texttt{-{}-enable-mis -{}-attention-backend flashinfer -{}-linear-attn-prefill-backend triton}, which turns CUDA graphs, the radix cache and chunked prefill off \\
vllm-jev & 0.3.7 on vLLM 0.29.0 & \texttt{vllm-jev serve} with its defaults, and with \texttt{-{}-radixflow true} \\
Flide & release & \texttt{serve} with two front-end processes and default batching \\
\bottomrule
\end{tabular}
\caption{Engine versions and launch settings.}
\label{tab:launch}
\end{table*}

%% file: tables/full_q4b.tex
\begin{table*}[t]
\centering
\tabsetup
\fontsize{9}{11}\selectfont\setlength{\tabcolsep}{3pt}
\begin{tabular}{l@{\hspace{\tabcolsep}\hspace{0pt}\hspace{\tabcolsep}}l@{\hspace{\tabcolsep}\hspace{0pt}\hspace{\tabcolsep}}rrr@{\hspace{\tabcolsep}\hspace{0pt}\hspace{\tabcolsep}}rrr@{\hspace{\tabcolsep}\hspace{0pt}\hspace{\tabcolsep}}rrr@{\hspace{\tabcolsep}\hspace{0pt}\hspace{\tabcolsep}}rrr}
\toprule
& & \multicolumn{3}{c}{\textbf{1 client}} & \multicolumn{3}{c}{\textbf{8 clients}} & \multicolumn{3}{c}{\textbf{32 clients}} & \multicolumn{3}{c}{\textbf{64 clients}} \\
\cmidrule(lr){3-5}\cmidrule(lr){6-8}\cmidrule(lr){9-11}\cmidrule(lr){12-14}
\textbf{Workload} & \textbf{Engine} & Med.$\downarrow$ & P99$\downarrow$ & Thr.$\uparrow$ & Med.$\downarrow$ & P99$\downarrow$ & Thr.$\uparrow$ & Med.$\downarrow$ & P99$\downarrow$ & Thr.$\uparrow$ & Med.$\downarrow$ & P99$\downarrow$ & Thr.$\uparrow$ \\
\midrule
\multirow{3}{*}{RACE, one} & vLLM & 24.6 & 37.6 & 38.5 & 59.9 & 80.6 & 130 & 164 & 194 & 192 & 315 & 428 & 196 \\
 & SGLang & 31.6 & 46.8 & 30.1 & 83.4 & 99.2 & 96.5 & 182 & 216 & 173 & 351 & 440 & 179 \\
 & Flide & \textbf{10.9} & \textbf{20.1} & \textbf{88.8} & \textbf{45.2} & \textbf{58.3} & \textbf{175} & \textbf{155} & \textbf{191} & \textbf{201} & \textbf{308} & \textbf{393} & \textbf{204} \\
\hdashline\noalign{\vskip 1.5pt}
\multirow{3}{*}{QuALITY, one} & vLLM & 68.6 & 79.1 & 14.7 & 395 & 699 & 19.3 & 1.7k & 1.8k & 19.3 & 3.3k & \textbf{3.4k} & 19.3 \\
 & SGLang & 75.3 & 85.6 & 13.6 & \textbf{386} & 523 & 18.5 & 1.7k & 1.8k & 18.5 & 3.4k & 3.6k & 18.5 \\
 & Flide & \textbf{62.3} & \textbf{68.5} & \textbf{16.6} & 413 & \textbf{437} & \textbf{19.4} & \textbf{1.6k} & \textbf{1.7k} & \textbf{19.9} & \textbf{3.2k} & \textbf{3.4k} & \textbf{19.9} \\
\hdashline\noalign{\vskip 1.5pt}
\multirow{4}{*}{DI} & vLLM & 27.7 & 492 & 15.3 & 180 & 898 & 31.6 & 771 & 2.1k & 35.9 & 1.6k & 4.2k & 36.0 \\
 & SGLang & 34.8 & 498 & 13.7 & 260 & 1.1k & 24.0 & 1.2k & 2.1k & 27.2 & 2.2k & 4.4k & 27.9 \\
 & MIS & 67.3 & 352 & 12.1 & 308 & 1.1k & 20.2 & 974 & 2.1k & 30.7 & 1.8k & 3.0k & 33.2 \\
 & Flide & \textbf{11.9} & \textbf{268} & \textbf{37.4} & \textbf{139} & \textbf{559} & \textbf{48.3} & \textbf{614} & \textbf{1.4k} & \textbf{50.2} & \textbf{1.3k} & \textbf{2.0k} & \textbf{50.4} \\
\hdashline\noalign{\vskip 1.5pt}
\multirow{4}{*}{RACE, all} & vLLM & 57.6 & 87.4 & 17.0 & 157 & 269 & 49.2 & 490 & 657 & 65.6 & 970 & 1.2k & 67.0 \\
 & SGLang & 37.4 & 55.4 & 25.5 & 173 & 222 & 45.4 & 640 & 1.1k & 48.3 & 1.3k & 1.4k & 49.8 \\
 & MIS & 66.6 & 69.3 & 14.9 & 160 & 412 & 48.6 & 407 & 695 & 67.0 & 892 & 1.2k & 75.6 \\
 & Flide & \textbf{14.1} & \textbf{23.9} & \textbf{70.2} & \textbf{68.8} & \textbf{84.8} & \textbf{116} & \textbf{258} & \textbf{371} & \textbf{122} & \textbf{505} & \textbf{565} & \textbf{124} \\
\hdashline\noalign{\vskip 1.5pt}
\multirow{4}{*}{QuALITY, all} & vLLM & 476 & 642 & 2.2 & 1.8k & 3.1k & 4.3 & 6.8k & 13k & 4.4 & 13k & 27k & 4.2 \\
 & SGLang & 727 & 1.3k & 1.4 & 3.4k & 4.3k & 2.4 & 13k & 17k & 2.4 & 27k & 38k & 2.1 \\
 & MIS & 166 & 206 & 6.4 & 1.2k & 1.5k & 7.0 & 4.8k & 5.6k & 6.7 & 8.8k & 10k & 6.9 \\
 & Flide & \textbf{141} & \textbf{170} & \textbf{8.0} & \textbf{963} & \textbf{1.2k} & \textbf{8.6} & \textbf{3.6k} & \textbf{4.2k} & \textbf{8.6} & \textbf{7.2k} & \textbf{8.1k} & \textbf{8.6} \\
\hdashline\noalign{\vskip 1.5pt}
\multirow{4}{*}{MultiRC} & vLLM & 281 & 492 & 3.7 & 1.1k & 1.8k & 7.1 & 4.3k & 5.7k & 7.3 & 13k & 15k & 4.9 \\
 & SGLang & 272 & 487 & 3.5 & 1.5k & 2.0k & 5.2 & 5.8k & 6.8k & 5.4 & 12k & 13k & 5.3 \\
 & MIS & 86.1 & 370 & 11.5 & 725 & 1.1k & 10.5 & 2.4k & 3.3k & 13.0 & 3.9k & 4.7k & 15.6 \\
 & Flide & \textbf{50.3} & \textbf{65.0} & \textbf{21.4} & \textbf{316} & \textbf{388} & \textbf{25.5} & \textbf{1.3k} & \textbf{1.4k} & \textbf{25.9} & \textbf{2.4k} & \textbf{2.6k} & \textbf{25.9} \\
\bottomrule
\end{tabular}
\caption{Full results on Qwen3.5 4B. Median and 99th percentile latency in ms (k for thousands) and requests per second at every number of clients.}
\label{tab:fullq4b}
\end{table*}

%% file: tables/full_ge4b.tex
\begin{table*}[t]
\centering
\tabsetup
\fontsize{9}{11}\selectfont\setlength{\tabcolsep}{3pt}
\begin{tabular}{l@{\hspace{\tabcolsep}\hspace{\dimexpr(\textwidth-458.0pt)/5\relax}\hspace{\tabcolsep}}l@{\hspace{\tabcolsep}\hspace{\dimexpr(\textwidth-458.0pt)/5\relax}\hspace{\tabcolsep}}rrr@{\hspace{\tabcolsep}\hspace{\dimexpr(\textwidth-458.0pt)/5\relax}\hspace{\tabcolsep}}rrr@{\hspace{\tabcolsep}\hspace{\dimexpr(\textwidth-458.0pt)/5\relax}\hspace{\tabcolsep}}rrr@{\hspace{\tabcolsep}\hspace{\dimexpr(\textwidth-458.0pt)/5\relax}\hspace{\tabcolsep}}rrr}
\toprule
& & \multicolumn{3}{c}{\textbf{1 client}} & \multicolumn{3}{c}{\textbf{8 clients}} & \multicolumn{3}{c}{\textbf{32 clients}} & \multicolumn{3}{c}{\textbf{64 clients}} \\
\cmidrule(lr){3-5}\cmidrule(lr){6-8}\cmidrule(lr){9-11}\cmidrule(lr){12-14}
\textbf{Workload} & \textbf{Engine} & Med.$\downarrow$ & P99$\downarrow$ & Thr.$\uparrow$ & Med.$\downarrow$ & P99$\downarrow$ & Thr.$\uparrow$ & Med.$\downarrow$ & P99$\downarrow$ & Thr.$\uparrow$ & Med.$\downarrow$ & P99$\downarrow$ & Thr.$\uparrow$ \\
\midrule
\multirow{3}{*}{RACE, one} & vLLM & 14.5 & 23.9 & 67.1 & 51.1 & 66.5 & 154 & \textbf{167} & 212 & 187 & \textbf{325} & 379 & 192 \\
 & SGLang & 36.6 & 45.7 & 27.1 & 76.0 & 89.4 & 105 & 203 & 278 & 155 & 361 & 474 & 173 \\
 & Flide & \textbf{11.3} & \textbf{17.2} & \textbf{87.5} & \textbf{45.9} & \textbf{56.0} & \textbf{172} & 169 & \textbf{193} & \textbf{188} & 327 & \textbf{362} & \textbf{193} \\
\hdashline\noalign{\vskip 1.5pt}
\multirow{3}{*}{QuALITY, one} & vLLM & 65.7 & 71.4 & 15.6 & \textbf{396} & 532 & \textbf{18.0} & \textbf{1.7k} & \textbf{1.8k} & \textbf{18.0} & \textbf{3.5k} & \textbf{3.7k} & \textbf{18.0} \\
 & SGLang & 95.3 & 103 & 10.9 & 543 & 727 & 13.2 & 2.3k & 2.5k & 13.2 & 4.9k & 5.0k & 13.2 \\
 & Flide & \textbf{63.8} & \textbf{68.8} & \textbf{16.1} & 452 & \textbf{474} & 17.8 & 1.8k & 1.9k & 17.9 & 3.6k & \textbf{3.7k} & 17.9 \\
\hdashline\noalign{\vskip 1.5pt}
\multirow{3}{*}{DI} & vLLM & 14.8 & 337 & 26.3 & \textbf{125} & 920 & 39.7 & 745 & 2.0k & 42.8 & 1.4k & 3.5k & 42.6 \\
 & SGLang & 38.3 & 593 & 12.7 & 238 & 1.2k & 23.1 & 1.1k & 2.2k & 26.9 & 2.3k & 4.5k & 26.7 \\
 & Flide & \textbf{11.6} & \textbf{284} & \textbf{36.1} & 142 & \textbf{693} & \textbf{45.5} & \textbf{626} & \textbf{1.4k} & \textbf{47.1} & \textbf{1.3k} & \textbf{2.2k} & \textbf{47.1} \\
\hdashline\noalign{\vskip 1.5pt}
\multirow{3}{*}{RACE, all} & vLLM & 25.8 & 40.3 & 38.2 & 81.2 & 131 & 95.4 & 293 & 437 & 107 & 592 & 738 & 109 \\
 & SGLang & 41.5 & 56.8 & 23.4 & 190 & 259 & 37.7 & 722 & 869 & 45.6 & 1.4k & 1.6k & 45.4 \\
 & Flide & \textbf{15.3} & \textbf{20.0} & \textbf{65.8} & \textbf{74.0} & \textbf{92.0} & \textbf{106} & \textbf{276} & \textbf{336} & \textbf{115} & \textbf{540} & \textbf{594} & \textbf{117} \\
\hdashline\noalign{\vskip 1.5pt}
\multirow{3}{*}{QuALITY, all} & vLLM & 364 & 493 & 3.0 & 1.9k & 2.4k & 4.3 & 7.2k & 8.5k & 4.3 & 15k & 18k & 4.2 \\
 & SGLang & 708 & 998 & 1.4 & 4.2k & 5.1k & 1.9 & 16k & 21k & 1.9 & 32k & 42k & 1.8 \\
 & Flide & \textbf{148} & \textbf{187} & \textbf{7.3} & \textbf{1.1k} & \textbf{1.3k} & \textbf{7.8} & \textbf{4.1k} & \textbf{4.7k} & \textbf{7.8} & \textbf{8.0k} & \textbf{9.0k} & \textbf{7.8} \\
\hdashline\noalign{\vskip 1.5pt}
\multirow{3}{*}{MultiRC} & vLLM & 87.2 & 140 & 11.8 & 338 & 488 & 23.3 & 1.3k & 1.5k & 24.3 & 2.6k & 3.0k & 23.8 \\
 & SGLang & 265 & 430 & 3.6 & 1.5k & 1.8k & 5.6 & 5.8k & 6.5k & 5.5 & 11k & 13k & 5.5 \\
 & Flide & \textbf{50.2} & \textbf{63.6} & \textbf{21.7} & \textbf{318} & \textbf{381} & \textbf{25.3} & \textbf{1.2k} & \textbf{1.4k} & \textbf{25.8} & \textbf{2.4k} & \textbf{2.7k} & \textbf{25.7} \\
\bottomrule
\end{tabular}
\caption{Full results on Gemma 4 E4B. Median and 99th percentile latency in ms (k for thousands) and requests per second at every number of clients.}
\label{tab:fullge4b}
\end{table*}

%% file: tables/exact.tex
\begin{table*}[t]
\centering
\tabsetup
\fontsize{9}{11}\selectfont\setlength{\tabcolsep}{5pt}
\begin{tabular}{l@{\hspace{\tabcolsep}\hspace{\dimexpr(\textwidth-444.7pt)/5\relax}\hspace{\tabcolsep}}l@{\hspace{\tabcolsep}\hspace{\dimexpr(\textwidth-444.7pt)/5\relax}\hspace{\tabcolsep}}rrr@{\hspace{\tabcolsep}\hspace{\dimexpr(\textwidth-444.7pt)/5\relax}\hspace{\tabcolsep}}rrr@{\hspace{\tabcolsep}\hspace{\dimexpr(\textwidth-444.7pt)/5\relax}\hspace{\tabcolsep}}rrr@{\hspace{\tabcolsep}\hspace{\dimexpr(\textwidth-444.7pt)/5\relax}\hspace{\tabcolsep}}rrr}
\toprule
& & \multicolumn{3}{c}{\textbf{DI}} & \multicolumn{3}{c}{\textbf{RACE}} & \multicolumn{3}{c}{\textbf{QuALITY}} & \multicolumn{3}{c}{\textbf{MultiRC}} \\
\cmidrule(lr){3-5}\cmidrule(lr){6-8}\cmidrule(lr){9-11}\cmidrule(lr){12-14}
\textbf{Qwen3.5} & \textbf{Engine} & TV$\downarrow$ & P99$\downarrow$ & Flip$\downarrow$ & TV$\downarrow$ & P99$\downarrow$ & Flip$\downarrow$ & TV$\downarrow$ & P99$\downarrow$ & Flip$\downarrow$ & TV$\downarrow$ & P99$\downarrow$ & Flip$\downarrow$ \\
\midrule
\multirow{4}{*}{0.8B} & vLLM & 1.05 & 3.3 & 2.3 & 1.08 & 4.3 & \textbf{1.4} & 1.34 & 4.3 & 3.2 & 0.58 & 2.7 & 0.3 \\
 & SGLang & 1.09 & 3.3 & 3.2 & 1.07 & 4.2 & \textbf{1.4} & 1.41 & 4.5 & 2.6 & 0.83 & 3.1 & 1.1 \\
 & MIS & 7.47 & 40.2 & 12.1 & 9.16 & 32.3 & 14.1 & 14.80 & 42.3 & 28.9 & 3.10 & 16.1 & 1.7 \\
 & Flide & \textbf{0.67} & \textbf{2.2} & \textbf{1.0} & \textbf{0.91} & \textbf{3.7} & 2.3 & \textbf{1.11} & \textbf{3.9} & \textbf{2.1} & \textbf{0.37} & \textbf{1.9} & \textbf{0.2} \\
\hdashline\noalign{\vskip 1.5pt}
\multirow{4}{*}{4B} & vLLM & 0.73 & 3.2 & 1.0 & 0.28 & 2.9 & \textbf{0.3} & 0.56 & 4.0 & \textbf{0.9} & 0.59 & 3.0 & 0.8 \\
 & SGLang & 0.75 & 3.5 & 1.0 & 0.29 & 3.1 & 1.3 & 0.60 & 4.3 & \textbf{0.9} & 0.85 & 3.0 & 1.7 \\
 & MIS & 5.77 & 25.4 & 6.3 & 2.51 & 28.8 & 3.1 & 5.73 & 42.3 & 8.1 & 4.69 & 29.8 & 5.3 \\
 & Flide & \textbf{0.51} & \textbf{2.5} & \textbf{0.8} & \textbf{0.23} & \textbf{2.4} & 0.7 & \textbf{0.45} & \textbf{3.4} & \textbf{0.9} & \textbf{0.36} & \textbf{1.9} & \textbf{0.5} \\
\hdashline\noalign{\vskip 1.5pt}
\multirow{4}{*}{9B} & vLLM & 0.46 & 3.2 & 1.1 & 0.27 & 2.5 & 0.4 & 0.55 & 3.1 & 0.8 & 0.61 & 2.8 & 1.5 \\
 & SGLang & 0.47 & 3.3 & 0.9 & 0.30 & 2.6 & \textbf{0.2} & 0.56 & 3.2 & \textbf{0.6} & 0.61 & 2.8 & 1.0 \\
 & MIS & 3.47 & 34.4 & 4.4 & 2.04 & 19.6 & 2.4 & 3.87 & 25.3 & 5.7 & 4.79 & 24.1 & 4.7 \\
 & Flide & \textbf{0.32} & \textbf{2.7} & \textbf{0.6} & \textbf{0.21} & \textbf{1.9} & 0.3 & \textbf{0.40} & \textbf{2.5} & \textbf{0.6} & \textbf{0.32} & \textbf{1.5} & \textbf{0.5} \\
\hdashline\noalign{\vskip 1.5pt}
\multirow{4}{*}{27B} & vLLM & 0.33 & 2.9 & 0.4 & 0.12 & 2.1 & \textbf{0.0} & 0.32 & 2.7 & 0.4 & 0.23 & 2.8 & 0.3 \\
 & SGLang & 0.34 & 3.2 & 0.7 & 0.14 & 2.3 & 0.2 & 0.35 & 2.9 & 0.5 & 0.29 & 3.3 & 0.3 \\
 & MIS & 2.65 & 24.6 & 3.3 & 0.87 & 14.5 & 0.8 & 2.34 & 23.1 & 3.1 & 2.50 & 30.1 & 2.1 \\
 & Flide & \textbf{0.24} & \textbf{2.6} & \textbf{0.3} & \textbf{0.10} & \textbf{1.6} & \textbf{0.0} & \textbf{0.24} & \textbf{2.2} & \textbf{0.2} & \textbf{0.18} & \textbf{2.0} & \textbf{0.1} \\
\midrule
\multirow{4}{*}{Average} & vLLM & 0.64 & 3.2 & 1.2 & 0.44 & 3.0 & \textbf{0.5} & 0.69 & 3.5 & 1.3 & 0.50 & 2.8 & 0.7 \\
 & SGLang & 0.66 & 3.3 & 1.4 & 0.45 & 3.0 & 0.8 & 0.73 & 3.7 & 1.2 & 0.65 & 3.1 & 1.0 \\
 & MIS & 4.84 & 31.2 & 6.5 & 3.64 & 23.8 & 5.1 & 6.69 & 33.2 & 11.4 & 3.77 & 25.1 & 3.5 \\
 & Flide & \textbf{0.44} & \textbf{2.5} & \textbf{0.7} & \textbf{0.36} & \textbf{2.4} & 0.8 & \textbf{0.55} & \textbf{3.0} & \textbf{1.0} & \textbf{0.30} & \textbf{1.8} & \textbf{0.3} \\
\bottomrule
\end{tabular}
\caption{Multi-item scoring moves the answers it shares, with every question of a state in one request.}
\label{tab:exact}
\end{table*}

%% file: tables/det.tex
\begin{table*}[t]
\centering
\tabsetup
\fontsize{9}{11}\selectfont\setlength{\tabcolsep}{1.5pt}
\begin{tabular}{l@{\hspace{\tabcolsep}\hspace{\dimexpr(\textwidth-467.7pt)/7\relax}\hspace{\tabcolsep}}l@{\hspace{\tabcolsep}\hspace{\dimexpr(\textwidth-467.7pt)/7\relax}\hspace{\tabcolsep}}rrr@{\hspace{\tabcolsep}\hspace{\dimexpr(\textwidth-467.7pt)/7\relax}\hspace{\tabcolsep}}rrr@{\hspace{\tabcolsep}\hspace{\dimexpr(\textwidth-467.7pt)/7\relax}\hspace{\tabcolsep}}rrr@{\hspace{\tabcolsep}\hspace{\dimexpr(\textwidth-467.7pt)/7\relax}\hspace{\tabcolsep}}rrr@{\hspace{\tabcolsep}\hspace{\dimexpr(\textwidth-467.7pt)/7\relax}\hspace{\tabcolsep}}rrr@{\hspace{\tabcolsep}\hspace{\dimexpr(\textwidth-467.7pt)/7\relax}\hspace{\tabcolsep}}rrr}
\toprule
& & \multicolumn{6}{c}{\textbf{One question per state}} & \multicolumn{12}{c}{\textbf{All questions of a state}} \\
\cmidrule(lr){3-8}\cmidrule(lr){9-20}
& & \multicolumn{3}{c}{RACE} & \multicolumn{3}{c}{QuALITY} & \multicolumn{3}{c}{DI} & \multicolumn{3}{c}{RACE} & \multicolumn{3}{c}{QuALITY} & \multicolumn{3}{c}{MultiRC} \\
\cmidrule(lr){3-5}\cmidrule(lr){6-8}\cmidrule(lr){9-11}\cmidrule(lr){12-14}\cmidrule(lr){15-17}\cmidrule(lr){18-20}
\textbf{Model} & \textbf{Mode} & Lat.$\downarrow$ & C64$\uparrow$ & Bits$\uparrow$ & Lat.$\downarrow$ & C64$\uparrow$ & Bits$\uparrow$ & Lat.$\downarrow$ & C64$\uparrow$ & Bits$\uparrow$ & Lat.$\downarrow$ & C64$\uparrow$ & Bits$\uparrow$ & Lat.$\downarrow$ & C64$\uparrow$ & Bits$\uparrow$ & Lat.$\downarrow$ & C64$\uparrow$ & Bits$\uparrow$ \\
\midrule
\multicolumn{20}{l}{\textbf{Qwen3.5}} \\
\multirow{2}{*}{0.8B} & Default & 6.1 & 654 & 88.8 & 17.9 & 96.7 & 99.5 & 6.8 & 237 & 85.5 & 7.4 & 544 & 73.4 & 37.3 & 38.6 & 98.2 & 17.2 & 121 & 94.1 \\
 & Invariant & 7.7 & 754 & 100 & 18.3 & 94.4 & 100 & 8.0 & 229 & 100 & 8.9 & 505 & 100 & 38.8 & 36.9 & 100 & 17.8 & 115 & 100 \\
\hdashline\noalign{\vskip 1.5pt}
\multirow{2}{*}{4B} & Default & 10.8 & 204 & 26.7 & 62.4 & 19.9 & 86.4 & 11.7 & 50.4 & 67.1 & 14.3 & 118 & 33.2 & 142 & 8.6 & 92.6 & 50.7 & 25.6 & 31.4 \\
 & Invariant & 14.5 & 184 & 100 & 63.0 & 19.6 & 100 & 15.4 & 49.0 & 100 & 16.7 & 118 & 100 & 146 & 8.3 & 100 & 51.3 & 24.9 & 100 \\
\hdashline\noalign{\vskip 1.5pt}
\multirow{2}{*}{9B} & Default & 14.2 & 116 & 27.2 & 103 & 11.3 & 90.9 & 15.1 & 28.7 & 60.0 & 20.7 & 70.3 & 35.7 & 235 & 5.0 & 76.4 & 83.4 & 14.5 & 44.0 \\
 & Invariant & 19.2 & 109 & 100 & 104 & 11.2 & 100 & 20.4 & 28.2 & 100 & 22.8 & 67.2 & 100 & 241 & 4.9 & 100 & 83.4 & 14.3 & 100 \\
\midrule
\multicolumn{20}{l}{\textbf{Gemma 4}} \\
\multirow{2}{*}{E2B} & Default & 8.6 & 354 & 13.8 & 37.2 & 32.2 & 100 & 9.0 & 86.6 & 88.6 & 10.5 & 218 & 61.9 & 87.4 & 13.5 & 100 & 28.2 & 47.2 & 100 \\
 & Invariant & 13.5 & 331 & 100 & 40.5 & 31.6 & 100 & 13.9 & 84.3 & 100 & 14.7 & 206 & 100 & 92.0 & 12.7 & 100 & 31.0 & 43.6 & 100 \\
\hdashline\noalign{\vskip 1.5pt}
\multirow{2}{*}{E4B} & Default & 11.2 & 190 & 84.9 & 64.7 & 17.9 & 100 & 11.8 & 46.8 & 92.2 & 15.0 & 117 & 100 & 147 & 7.8 & 100 & 49.8 & 25.8 & 100 \\
 & Invariant & 17.5 & 181 & 100 & 68.4 & 17.6 & 100 & 17.8 & 45.9 & 100 & 20.5 & 112 & 100 & 154 & 7.5 & 100 & 53.1 & 24.3 & 100 \\
\hdashline\noalign{\vskip 1.5pt}
\multirow{2}{*}{12B} & Default & 19.6 & 75.5 & 65.3 & 158 & 7.0 & 100 & 19.0 & 18.2 & 90.6 & 29.0 & 45.6 & 84.1 & 370 & 3.1 & 100 & 124 & 9.6 & 100 \\
 & Invariant & 25.2 & 72.5 & 100 & 163 & 6.9 & 100 & 27.7 & 18.0 & 100 & 33.7 & 44.0 & 100 & 388 & 2.9 & 100 & 129 & 9.0 & 100 \\
\bottomrule
\end{tabular}
\caption{Identical bits at any load. Bits gives the percent of answers whose bits do not change with the load.}
\label{tab:det}
\end{table*}

%% file: tables/split.tex
\begin{table}[t]
\centering
\tabsetup
\fontsize{9}{11}\selectfont\setlength{\tabcolsep}{2.4pt}
\begin{tabular}{l rr rr rr}
\toprule
& \multicolumn{2}{c}{\textbf{DI}} & \multicolumn{2}{c}{\textbf{QuALITY, all}} & \multicolumn{2}{c}{\textbf{MultiRC}} \\
\cmidrule(lr){2-3}\cmidrule(lr){4-5}\cmidrule(lr){6-7}
& Lat.$\downarrow$ & Tok.$\downarrow$ & Lat.$\downarrow$ & Tok.$\downarrow$ & Lat.$\downarrow$ & Tok.$\downarrow$ \\
\midrule
vLLM, endpoint & 26.7 & 2098 & 537 & 12448 & 154 & 17314 \\
vLLM, token ids & 20.1 & 2049 & 350 & 12448 & 97.3 & 17314 \\
\hdashline\noalign{\vskip 1.5pt}
SGLang, endpoint & 28.6 & 3268 & 359 & 30826 & 130 & 13358 \\
SGLang, token ids & 26.4 & 2574 & 155 & 30826 & 86.2 & 13625 \\
MIS & 53.1 & 1364 & 66.2 & 7479 & 55.2 & 2717 \\
\hdashline\noalign{\vskip 1.5pt}
Flide & 6.6 & 1357 & 38.1 & 7460 & 17.7 & 2668 \\
\bottomrule
\end{tabular}
\caption{Where the time goes. One request alone on 0.8B, ms and tokens computed.}
\label{tab:split}
\end{table}

%% file: tables/resources.tex
\begin{table*}[t]
\centering
\tabsetup
\fontsize{9}{11}\selectfont\setlength{\tabcolsep}{3.6pt}
\begin{tabular}{l rr rrrr rrrr rrrr}
\toprule
& \multicolumn{2}{c}{\textbf{Memory, GiB}} & \multicolumn{4}{c}{\textbf{DI}} & \multicolumn{4}{c}{\textbf{RACE}} & \multicolumn{4}{c}{\textbf{QuALITY, all}} \\
\cmidrule(lr){2-3}\cmidrule(lr){4-7}\cmidrule(lr){8-11}\cmidrule(lr){12-15}
& & & \multicolumn{2}{c}{1 client} & \multicolumn{2}{c}{64 clients} & \multicolumn{2}{c}{1 client} & \multicolumn{2}{c}{64 clients} & \multicolumn{2}{c}{1 client} & \multicolumn{2}{c}{64 clients} \\
\textbf{Engine} & Peak$\downarrow$ & Graphs$\downarrow$ & Busy & Tok.$\uparrow$ & Busy & Tok.$\uparrow$ & Busy & Tok.$\uparrow$ & Busy & Tok.$\uparrow$ & Busy & Tok.$\uparrow$ & Busy & Tok.$\uparrow$ \\
\midrule
vLLM & 135.3 & 0.6 & 50 & 17.9 & 77 & 33.6 & 35 & 12.5 & 60 & 64.9 & 39 & 11.3 & 51 & 15.5 \\
SGLang & 131.7 & 1.9 & 59 & 19.9 & 91 & 37.1 & 28 & 10.4 & 60 & 60.5 & 67 & 10.5 & 84 & 16.5 \\
MIS & 129.0 & -- & 32 & 16.7 & 71 & 46.4 & 17 & 5.0 & 53 & 46.6 & 86 & 47.7 & 84 & 51.0 \\
\hdashline\noalign{\vskip 1.5pt}
vLLM, 10\% & \textbf{17.2} & \textbf{0.1} & 46 & 15.9 & 76 & 31.5 & 27 & 9.8 & 60 & 65.0 & 34 & 9.8 & 44 & 12.9 \\
SGLang, 10\% & 18.7 & 1.4 & 58 & 18.7 & 82 & 32.8 & 27 & 10.1 & 56 & 54.0 & 66 & 10.6 & 82 & 15.3 \\
\hdashline\noalign{\vskip 1.5pt}
Flide & 18.8 & 2.1 & 77 & \textbf{49.5} & 88 & \textbf{68.4} & 64 & \textbf{30.0} & 58 & \textbf{67.6} & 92 & \textbf{59.0} & 88 & \textbf{63.5} \\
\bottomrule
\end{tabular}
\caption{Memory and GPU use on Qwen3.5-4B. Busy is the percent of time a kernel runs, Tok. the useful tokens
per second in thousands, and the 10\% rows cap the engine at a tenth of the card.}
\label{tab:resources}
\end{table*}

%% file: tables/gqa.tex
\begin{table}[t]
\centering
\tabsetup
\fontsize{9}{11}\selectfont\setlength{\tabcolsep}{2.75pt}
\begin{tabular}{ll rrr rrr}
\toprule
& & \multicolumn{3}{c}{\textbf{Speed}} & \multicolumn{3}{c}{\textbf{Answers}} \\
\cmidrule(lr){3-5}\cmidrule(lr){6-8}
\textbf{Qwen3.5} & \textbf{Engine} & Lat.$\downarrow$ & C8$\uparrow$ & C64$\uparrow$ & TV$\downarrow$ & Flip$\downarrow$ & Acc.$\uparrow$ \\
\midrule
\multirow{3}{*}{0.8B} & vLLM & 168.0 & 16.8 & 22.8 & 1.06 & 1.4 & 70.1 \\
 & SGLang & 251.6 & 7.6 & 8.3 & 1.05 & 1.4 & 70.0 \\
 & Flide & \textbf{22.4} & \textbf{88.9} & \textbf{123.1} & \textbf{0.82} & \textbf{0.7} & \textbf{70.7} \\
\hdashline\noalign{\vskip 1.5pt}
\multirow{3}{*}{4B} & vLLM & 194.0 & 13.3 & 16.8 & 0.70 & 0.8 & 77.7 \\
 & SGLang & 282.8 & 7.1 & 8.3 & 0.74 & 1.1 & \textbf{77.9} \\
 & Flide & \textbf{31.2} & \textbf{50.0} & \textbf{52.5} & \textbf{0.51} & \textbf{0.5} & 77.6 \\
\hdashline\noalign{\vskip 1.5pt}
\multirow{3}{*}{9B} & vLLM & 198.3 & 9.8 & 10.0 & 0.73 & 1.2 & \textbf{79.3} \\
 & SGLang & 277.5 & 7.6 & 8.2 & 0.77 & 1.1 & 79.1 \\
 & Flide & \textbf{40.0} & \textbf{35.2} & \textbf{36.6} & \textbf{0.48} & \textbf{0.8} & 79.2 \\
\bottomrule
\end{tabular}
\caption{Questions about images. GQA, every question of an image in one Flide request.}
\label{tab:gqa}
\end{table}

%% file: tables/repeat.tex
\begin{table}[t]
\centering
\tabsetup
\fontsize{9}{11}\selectfont\setlength{\tabcolsep}{2.3pt}
\begin{tabular}{ll rr rr rr}
\toprule
& & \multicolumn{2}{c}{\textbf{DI}} & \multicolumn{2}{c}{\textbf{RACE}} & \multicolumn{2}{c}{\textbf{QuALITY}} \\
\cmidrule(lr){3-4}\cmidrule(lr){5-6}\cmidrule(lr){7-8}
\textbf{Qwen3.5} & \textbf{Engine} & Lat.$\downarrow$ & C64$\uparrow$ & Lat.$\downarrow$ & C64$\uparrow$ & Lat.$\downarrow$ & C64$\uparrow$ \\
\midrule
\multirow{3}{*}{0.8B} & vLLM & 24.8 & 41.0 & 24.1 & 403.9 & 36.0 & 123.1 \\
 & SGLang & 28.9 & 100.0 & 26.9 & 372.1 & 35.1 & 98.0 \\
 & Flide & \textbf{7.3} & \textbf{209.5} & \textbf{7.0} & \textbf{529.4} & \textbf{11.0} & \textbf{177.1} \\
\hdashline\noalign{\vskip 1.5pt}
\multirow{3}{*}{4B} & vLLM & 28.5 & 24.9 & 27.1 & 184.6 & 39.1 & 38.8 \\
 & SGLang & 34.6 & 28.7 & 34.5 & 176.6 & 75.0 & 25.5 \\
 & Flide & \textbf{12.2} & \textbf{47.3} & \textbf{12.0} & \textbf{193.5} & \textbf{18.4} & \textbf{39.9} \\
\hdashline\noalign{\vskip 1.5pt}
\multirow{3}{*}{9B} & vLLM & 29.5 & 16.4 & 29.5 & 112.0 & 40.9 & 23.1 \\
 & SGLang & 35.3 & 17.1 & 33.7 & 110.2 & 113.6 & 15.6 \\
 & Flide & \textbf{15.6} & \textbf{27.7} & \textbf{15.0} & \textbf{116.8} & \textbf{22.1} & \textbf{23.7} \\
\bottomrule
\end{tabular}
\caption{States that recur. Each engine keeps states across requests with its own cache.}
\label{tab:repeat}
\end{table}

%% file: tables/startup.tex
\begin{table}[t]
\centering
\tabsetup
\fontsize{9}{11}\selectfont\setlength{\tabcolsep}{6pt}
\begin{tabular}{l rrr}
\toprule
\textbf{Start with} & Ready (s)$\downarrow$ & Compiled$\downarrow$ & Tuned$\downarrow$ \\
\midrule
Bundle & 36.3 & 0 & 0 \\
Bundle, other install & 33.3 & 0 & 0 \\
Copied cache & 181.0 & 231 & 200 \\
Empty cache & 180.6 & 231 & 200 \\
\bottomrule
\end{tabular}
\caption{Kernels shipped with the package. Start of a 0.8B server, kernels compiled and tuned.}
\label{tab:startup}
\end{table}

%% file: tables/models.tex
\begin{table*}[t]
\centering
\tabsetup
\fontsize{9}{11}\selectfont\setlength{\tabcolsep}{4pt}
\begin{tabular}{lll rrrrr}
\toprule
& & & \multicolumn{5}{c}{Speedup$\uparrow$} \\
\cmidrule(lr){4-8}
\textbf{Model} & \textbf{Backbone} & \textbf{Release} & DI & RACE & MultiRC & QuALITY & GQA \\
\midrule
\multicolumn{8}{l}{\textit{Fork}} \\
\href{https://huggingface.co/internlm/Intern-Decision-4B}{Intern-Decision 4B} & Qwen3.5 4B & transformers & 4.40 & 4.35 & 1.15 & 1.46 & 1.47 \\
\href{https://huggingface.co/WangSenQWQ/Senova-4B}{Senova 4B} & Qwen3.5 4B & SGLang & 3.13 & 3.65 & 28.10 & 1.18 &  \\
\href{https://huggingface.co/chinhnc/Quyet-1.0-Medium}{Quyet Medium} & Qwen3.5 4B & transformers & 5.95 & 5.90 & 7.02 & 1.37 &  \\
\href{https://huggingface.co/apertixai/cerno-4b}{cerno 4B} & Qwen3.5 4B & vLLM & 4.72 & 2.75 & 9.06 & 1.13 &  \\
\href{https://huggingface.co/nace-ai/drex-v1.5}{Drex v1.5 9B} & Qwen3.5 9B & transformers & 2.95 & 4.30 & 7.44 & 1.34 &  \\
\href{https://huggingface.co/eljiwo/jiwo-4b}{jiwo 4B} & Qwen3.5 4B & custom & 1.74 & 1.74 & 6.90 & 1.75 &  \\
\href{https://huggingface.co/jaredpalmer/kev-4b}{Kev 4B} & Qwen3.5 4B & custom & 5.60 & 1.51 & 1.30 & 1.15 &  \\
\href{https://huggingface.co/SimpleJev/JevAny-Qwen3.5-4B-LoRA}{JevAny 4B} & Qwen3.5 4B & custom & 1.48 & 1.10 & 5.85 & 1.17 & 4.39 \\
\href{https://huggingface.co/bespokelabs/Bespoke-Nimble-9B-v3}{Bespoke-Nimble 9B} & Qwen3.5 9B & transformers & 2.58 & 3.94 & 2.13 & 1.59 &  \\
\href{https://huggingface.co/torchcast-ai/torchcast-decision-27b}{torchcast 27B} & Qwen3.5 27B & transformers & 1.56 & 1.27 & 8.38 & 1.43 & 5.51 \\
\href{https://huggingface.co/FINAL-Bench/Darwin-27B-ZTC}{Darwin 27B} & Qwen3.5 27B & custom & 1.60 & 2.23 & 7.04 & 1.22 &  \\
\href{https://huggingface.co/FrontiersMind/Lumma-fev-27B}{Lumma 27B} & Qwen3.5 27B & custom & 1.74 & 2.17 & 7.14 & 1.22 & 5.37 \\
\href{https://huggingface.co/PNPM244/linne-fern-1-0}{Linne 27B} & Qwen3.5 27B & transformers & 2.33 & 2.57 & 8.55 & 1.22 &  \\
\href{https://huggingface.co/theunnecessarythings/JADE}{JADE 27B} & Qwen3.5 27B & vLLM & 2.00 & 2.33 & 9.17 & 1.75 &  \\
\href{https://huggingface.co/andyzhang232/ajev-gemma4-12b-lora7}{AJev 12B} & Gemma 4 12B & transformers & 3.71 & 4.29 & 9.30 & 2.60 &  \\
\href{https://huggingface.co/PelaAI/KnowLine-26B-A4B-Gen2.5}{KnowLine 26B} & Gemma 4 26B-A4B & SGLang & 1.59 & 2.07 & 5.79 & 1.77 &  \\
\href{https://huggingface.co/SkyPanther/synack-decide-26b-a4b}{SynACK Decide 26B} & Gemma 4 26B-A4B & custom & 3.33 & 3.21 & 11.87 & 3.02 &  \\
\href{https://huggingface.co/org2ai/Wald-26B-A4B}{Wald 26B} & Gemma 4 26B-A4B & vLLM & 1.56 & 1.85 & 12.71 & 1.54 &  \\
\href{https://huggingface.co/juspay/Xor-26B-A4B}{Xor 26B} & Gemma 4 26B-A4B & SGLang & 2.49 & 2.76 & 6.17 & 2.77 & 6.01 \\
\href{https://huggingface.co/sthanika-ai/Sieve-31B}{Sieve 31B} & Gemma 4 31B & transformers & 2.97 & 2.54 & 14.26 & 2.58 &  \\
\href{https://huggingface.co/moganai/mogan-decision-31B-it}{mogan 31B} & Gemma 4 31B & transformers & 1.57 & 1.38 & 8.53 & 1.78 &  \\
\href{https://huggingface.co/ZefanCai/Open-Jev-2B}{Open-Jev 2B} & Qwen3.5 2B & vllm-jev & 4.15 & 3.85 & 5.01 & 1.65 &  \\
\href{https://huggingface.co/OmniJev/OneJev1.5-0.8B}{OneJev 0.8B} & Qwen3.5 0.8B & qev & 11.89 &  & 7.28 & 4.71 &  \\
\hdashline\noalign{\vskip 1.5pt}
\multicolumn{8}{l}{\textit{Fork, two-way state}} \\
\href{https://huggingface.co/baselabs/elephant-1}{elephant 9B} & Qwen3.5 9B & transformers & 5.28 & 7.90 & 3.36 & 2.01 & 7.91 \\
\href{https://huggingface.co/baselabs/mammoth-1}{mammoth 27B} & Qwen3.5 27B & transformers & 3.29 & 4.86 & 2.26 & 1.62 & 5.95 \\
\hdashline\noalign{\vskip 1.5pt}
\multicolumn{8}{l}{\textit{Per question, two-way}} \\
\href{https://huggingface.co/tattrongvu/Jeffan-4B-2010}{Jeffan 4B} & Qwen3.5 4B & transformers & 3.13 & 5.33 & 2.62 & 1.87 &  \\
\href{https://huggingface.co/startlux-models/StartLux-Decision-v1.1-4B-Instant}{StartLux 4B} & Qwen3.5 4B & vLLM & 2.64 & 3.27 & 1.10 & 1.19 &  \\
\href{https://huggingface.co/LycheeAI/Lychee-27B}{Lychee 27B} & Qwen3.5 27B & transformers & 1.42 & 2.09 & 1.50 & 1.42 &  \\
\hdashline\noalign{\vskip 1.5pt}
\multicolumn{8}{l}{\textit{Per request, two-way}} \\
\href{https://huggingface.co/perplexity-ai/pplx-decider-flash-v1.2}{pplx-decider 4B} & Qwen3.5 4B & custom & 3.10 & 4.26 & 1.40 & 1.20 & 1.92 \\
\bottomrule
\end{tabular}
\caption{Decision models served by Flide. Median latency of each release engine over that of Flide for one request at a time, where custom marks an engine that the release ships itself.}
\label{tab:models}
\end{table*}

%% file: main.bbl
\begin{thebibliography}{55}
\providecommand{\natexlab}[1]{#1}
\providecommand{\url}[1]{\texttt{#1}}
\expandafter\ifx\csname urlstyle\endcsname\relax
  \providecommand{\doi}[1]{doi: #1}\else
  \providecommand{\doi}{doi: \begingroup \urlstyle{rm}\Url}\fi

\bibitem[Antunes et~al.(2026)Antunes, Albuquerque, Schmith, Batista, da~Silva,
  Rossi, da~Silva, da~Costa, Bertotto, and Fausto]{antunes2026ndm}
Carlos~Andr{\'e} Antunes, Aldry Albuquerque, Gabriel de~Melo Schmith, Nat{\~a}
  Batista, Pedro Oziel~Escobar da~Silva, Stefano~Benko Rossi, Lingrid
  Dietrich~Lorenz da~Silva, Fl{\'a}via~Cristina da~Costa, Thiago Bertotto, and
  Luis~Fernando Fausto.
\newblock Narrow decision models: A {System-1} architecture for calibrated,
  typed decisions.
\newblock \emph{arXiv preprint arXiv:2610.25553}, 2026.

\bibitem[Cai et~al.(2024)Cai, Li, Geng, Peng, Lee, Chen, and
  Dao]{cai2024medusa}
Tianle Cai, Yuhong Li, Zhengyang Geng, Hongwu Peng, Jason~D. Lee, Deming Chen,
  and Tri Dao.
\newblock Medusa: Simple {LLM} inference acceleration framework with multiple
  decoding heads.
\newblock In \emph{Proceedings of the 41st International Conference on Machine
  Learning}, pages 5209--5235, 2024.

\bibitem[Chouhan and Perry(2026)]{chouhan2026encode}
Avdhesh~Singh Chouhan and Jack Perry.
\newblock Verdict: Encode once, decide many.
\newblock \emph{arXiv preprint arXiv:2610.20778}, 2026.

\bibitem[{Decision Index}(2026)]{decisionindex2026}
{Decision Index}.
\newblock Jev decision index.
\newblock \url{https://huggingface.co/spaces/multimodalart/jev-decision-index},
  2026.

\bibitem[Du et~al.(2025)Du, Wang, Zhang, Cheng, Lan, Sang, Cheng, Yao,
  et~al.]{du2025prefillonly}
Kuntai Du, Bowen Wang, Chen Zhang, Yiming Cheng, Qing Lan, Hejian Sang, Yihua
  Cheng, Jiayi Yao, et~al.
\newblock {PrefillOnly}: An inference engine for prefill-only workloads in
  large language model applications.
\newblock \emph{arXiv preprint arXiv:2505.07203}, 2025.

\bibitem[{Gemma Team}(2026)]{gemmateam2026gemma4}
{Gemma Team}.
\newblock Gemma 4 technical report, 2026.
\newblock URL \url{https://arxiv.org/abs/2607.02770}.

\bibitem[Ghantasala(2026)]{ghantasala2026treewy}
Sneha~Murthy Ghantasala.
\newblock {TreeWY}: Speculative verification for gated {DeltaNet} hybrids.
\newblock \emph{arXiv preprint arXiv:2608.20961}, 2026.

\bibitem[Gu and Dao(2023)]{gu2023mamba}
Albert Gu and Tri Dao.
\newblock Mamba: Linear-time sequence modeling with selective state spaces.
\newblock \emph{arXiv preprint arXiv:2312.00752}, 2023.

\bibitem[He and {Thinking Machines Lab}(2025)]{he2025nondeterminism}
Horace He and {Thinking Machines Lab}.
\newblock Defeating nondeterminism in {LLM} inference.
\newblock
  \url{https://thinkingmachines.ai/blog/defeating-nondeterminism-in-llm-inference},
  2025.

\bibitem[Hudson and Manning(2019)]{hudson2019gqa}
Drew~A. Hudson and Christopher~D. Manning.
\newblock {GQA}: A new dataset for real-world visual reasoning and
  compositional question answering.
\newblock In \emph{Proceedings of the IEEE/CVF Conference on Computer Vision
  and Pattern Recognition}, pages 6700--6709, 2019.

\bibitem[Jin et~al.(2026)Jin, Ma, Yin, Chen, Song, Pang, and
  Zhang]{jin2026duplexjev}
Jie Jin, Ziyin Ma, Min Yin, Jinyu Chen, Haigang Song, Zhikun Pang, and Xiaowen
  Zhang.
\newblock Batched speech decisions without decoding: Single-token supervision
  lets a frozen {LLM} hear beyond the transcript.
\newblock \emph{arXiv preprint arXiv:2610.02638}, 2026.

\bibitem[Jing et~al.(2026)Jing, Hu, Chen, Shi, Huang, Zhang, Liu, Xie, and
  Li]{jing2026radixflow}
Huihao Jing, Wenbin Hu, Shaojin Chen, Haochen Shi, Haoyu Huang, Sirui Zhang,
  Yuxuan Liu, Zhongwei Xie, and Haoran Li.
\newblock {RadixFlow}: Filling execution bubbles in prefix-cached hybrid {LLM}
  serving.
\newblock \emph{arXiv preprint arXiv:2610.18143}, 2026.

\bibitem[Juravsky et~al.(2024)Juravsky, Brown, Ehrlich, Fu, R{\'e}, and
  Mirhoseini]{juravsky2024hydragen}
Jordan Juravsky, Bradley Brown, Ryan Ehrlich, Daniel~Y. Fu, Christopher R{\'e},
  and Azalia Mirhoseini.
\newblock Hydragen: High-throughput {LLM} inference with shared prefixes.
\newblock \emph{arXiv preprint arXiv:2402.05099}, 2024.

\bibitem[Khashabi et~al.(2018)Khashabi, Chaturvedi, Roth, Upadhyay, and
  Roth]{khashabi2018multirc}
Daniel Khashabi, Snigdha Chaturvedi, Michael Roth, Shyam Upadhyay, and Dan
  Roth.
\newblock Looking beyond the surface: A challenge set for reading comprehension
  over multiple sentences.
\newblock In \emph{Proceedings of the 2018 Conference of the North {A}merican
  Chapter of the Association for Computational Linguistics}, pages 252--262,
  2018.

\bibitem[Kwon et~al.(2023)Kwon, Li, Zhuang, Sheng, Zheng, Yu, Gonzalez, Zhang,
  and Stoica]{kwon2023vllm}
Woosuk Kwon, Zhuohan Li, Siyuan Zhuang, Ying Sheng, Lianmin Zheng, Cody~Hao Yu,
  Joseph~E. Gonzalez, Hao Zhang, and Ion Stoica.
\newblock Efficient memory management for large language model serving with
  {PagedAttention}.
\newblock In \emph{Proceedings of the 29th Symposium on Operating Systems
  Principles}, 2023.

\bibitem[Lai et~al.(2017)Lai, Xie, Liu, Yang, and Hovy]{lai2017race}
Guokun Lai, Qizhe Xie, Hanxiao Liu, Yiming Yang, and Eduard Hovy.
\newblock {RACE}: Large-scale {R}eading comprehension dataset from
  examinations.
\newblock In \emph{Proceedings of the 2017 Conference on Empirical Methods in
  Natural Language Processing}, 2017.

\bibitem[Ma et~al.(2026)Ma, Gao, Liu, Qiu, Li, Zhuang, Ji, and
  Sun]{ma2026fastjev}
Jie Ma, Jie Gao, Yihang Liu, Zhike Qiu, Junle Li, Chongyi Zhuang, Jiayi Ji, and
  Xiaoshuai Sun.
\newblock {FastJEV}: Understanding redundancy for compact {JEV} inference.
\newblock \emph{arXiv preprint arXiv:2610.11379}, 2026.

\bibitem[Ma(2026)]{ma2026lumotree}
Zhiyuan Ma.
\newblock {LumoTree}: Path-parallel speculative verification for hybrid
  language models.
\newblock \emph{arXiv preprint arXiv:2609.23900}, 2026.

\bibitem[Miao et~al.(2024)Miao, Oliaro, Zhang, Cheng, Wang, Zhang, Wong, Zhu,
  Yang, Shi, Shi, Chen, Arfeen, Abhyankar, and Jia]{miao2024specinfer}
Xupeng Miao, Gabriele Oliaro, Zhihao Zhang, Xinhao Cheng, Zeyu Wang, Zhengxin
  Zhang, Rae Ying~Yee Wong, Alan Zhu, Lijie Yang, Xiaoxiang Shi, Chunan Shi,
  Zhuoming Chen, Daiyaan Arfeen, Reyna Abhyankar, and Zhihao Jia.
\newblock {SpecInfer}: Accelerating large language model serving with
  tree-based speculative inference and verification.
\newblock In \emph{Proceedings of the 29th ACM International Conference on
  Architectural Support for Programming Languages and Operating Systems}, 2024.

\bibitem[{mode-io}(2026)]{modeio2026vllmjev}
{mode-io}.
\newblock {vLLM Jev}.
\newblock \url{https://github.com/mode-io/vllm-jev}, 2026.

\bibitem[{Nebius}(2024)]{nebius2024swetraj}
{Nebius}.
\newblock {SWE}-agent trajectories.
\newblock \url{https://huggingface.co/datasets/nebius/SWE-agent-trajectories},
  2024.

\bibitem[{OneJev Team}(2026)]{onejev2026}
{OneJev Team}.
\newblock {OneJev}, a multimodal system one decision model.
\newblock \url{https://github.com/OmniJev/OneJev}, 2026.

\bibitem[Ong et~al.(2024)Ong, Almahairi, Wu, Chiang, Wu, Gonzalez, Kadous, and
  Stoica]{ong2024routellm}
Isaac Ong, Amjad Almahairi, Vincent Wu, Wei-Lin Chiang, Tianhao Wu, Joseph~E.
  Gonzalez, M.~Waleed Kadous, and Ion Stoica.
\newblock {RouteLLM}: Learning to route {LLMs} with preference data.
\newblock \emph{arXiv preprint arXiv:2406.18665}, 2024.

\bibitem[{Open-Jev}(2026)]{openjev2026}
{Open-Jev}.
\newblock {Open-Jev-2B} and {Open-Jev-9B}.
\newblock \url{https://huggingface.co/ZefanCai/Open-Jev-2B},
  \url{https://huggingface.co/ZefanCai/Open-Jev-9B}, 2026.

\bibitem[Pan et~al.(2025)Pan, Wang, Jia, Karakus, Zancato, Dao, Wang, and
  Netravali]{pan2025marconi}
Rui Pan, Zhuang Wang, Zhen Jia, Can Karakus, Luca Zancato, Tri Dao, Yida Wang,
  and Ravi Netravali.
\newblock Marconi: Prefix caching for the era of hybrid {LLMs}.
\newblock In \emph{Proceedings of Machine Learning and Systems}, 2025.

\bibitem[Pang et~al.(2022)Pang, Parrish, Joshi, Nangia, Phang, Chen,
  Padmakumar, Ma, Thompson, He, and Bowman]{pang2022quality}
Richard~Yuanzhe Pang, Alicia Parrish, Nitish Joshi, Nikita Nangia, Jason Phang,
  Angelica Chen, Vishakh Padmakumar, Johnny Ma, Jana Thompson, He~He, and
  Samuel~R. Bowman.
\newblock {QuALITY}: Question answering with long input texts, yes!
\newblock In \emph{Proceedings of the 2022 Conference of the North American
  Chapter of the Association for Computational Linguistics}, 2022.

\bibitem[Paszke et~al.(2019)Paszke, Gross, Massa, Lerer, Bradbury, Chanan,
  Killeen, Lin, Gimelshein, Antiga, Desmaison, K{\"o}pf, Yang, DeVito, Raison,
  Tejani, Chilamkurthy, Steiner, Fang, Bai, and Chintala]{paszke2019pytorch}
Adam Paszke, Sam Gross, Francisco Massa, Adam Lerer, James Bradbury, Gregory
  Chanan, Trevor Killeen, Zeming Lin, Natalia Gimelshein, Luca Antiga, Alban
  Desmaison, Andreas K{\"o}pf, Edward Yang, Zachary DeVito, Martin Raison,
  Alykhan Tejani, Sasank Chilamkurthy, Benoit Steiner, Lu~Fang, Junjie Bai, and
  Soumith Chintala.
\newblock {PyTorch}: An imperative style, high-performance deep learning
  library.
\newblock In \emph{Advances in Neural Information Processing Systems},
  volume~32, 2019.

\bibitem[Patodiya(2026)]{patodiya2026samerequest}
Aditi Patodiya.
\newblock Same request, different answer: Quantization amplifies cache-induced
  divergence in {LLM} serving.
\newblock \emph{arXiv preprint arXiv:2609.04748}, 2026.

\bibitem[{Qwen Team}(2026)]{qwen2026qwen35}
{Qwen Team}.
\newblock {Qwen3.5}: Towards native multimodal agents.
\newblock \url{https://qwen.ai/blog?id=qwen3.5}, 2026.

\bibitem[Sellis(1988)]{sellis1988mqo}
Timos~K. Sellis.
\newblock Multiple-query optimization.
\newblock \emph{ACM Transactions on Database Systems}, 13\penalty0
  (1):\penalty0 23--52, 1988.

\bibitem[{SGLang Project}(2026{\natexlab{a}})]{sglang2026mis}
{SGLang Project}.
\newblock Fix {Qwen3.5} {GDN} multi-item scoring.
\newblock \url{https://github.com/sgl-project/sglang/pull/33922},
  2026{\natexlab{a}}.

\bibitem[{SGLang Project}(2026{\natexlab{b}})]{sglang2026misdecisions}
{SGLang Project}.
\newblock Decisions: Support cold shared-prefix prefill with {MIS}.
\newblock \url{https://github.com/sgl-project/sglang/pull/44793},
  2026{\natexlab{b}}.

\bibitem[{SGLang Project}(2026{\natexlab{c}})]{sglang2026systemone}
{SGLang Project}.
\newblock A system one compatible /v1/systemone route.
\newblock \url{https://github.com/sgl-project/sglang/pull/41208},
  2026{\natexlab{c}}.

\bibitem[{SGLang Team}(2025)]{sglang2025deterministic}
{SGLang Team}.
\newblock Towards deterministic inference in {SGLang} and reproducible {RL}
  training.
\newblock \url{https://lmsys.org/blog/2025-09-22-sglang-deterministic/}, 2025.

\bibitem[{SGLang Team}(2026)]{sglang2026cudagraph}
{SGLang Team}.
\newblock Advanced {CUDA} graph techniques in {SGLang}.
\newblock \url{https://www.lmsys.org/blog/2026-08-17-advanced-cuda-graph},
  2026.

\bibitem[Shirokikh and Nikolenko(2026)]{shirokikh2026sparse}
Mikhail Shirokikh and Sergey Nikolenko.
\newblock Sparse prefix caching for hybrid and recurrent {LLM} serving.
\newblock \emph{arXiv preprint arXiv:2605.05219}, 2026.

\bibitem[{StartLux}(2026)]{startlux2026decision}
{StartLux}.
\newblock {StartLux Decision} models.
\newblock \url{https://huggingface.co/startlux-models}, 2026.

\bibitem[{TypeSafe AI}(2026)]{typesafe2026jev}
{TypeSafe AI}.
\newblock Introducing system one models and {Jev}.
\newblock \url{https://typesafe.ai/blog/introducing-system-one-models-and-jev},
  2026.

\bibitem[{vLLM Project}(2026{\natexlab{a}})]{vllm2026gdninvariant}
{vLLM Project}.
\newblock {vLLM} v0.31.0 attention backend selector.
\newblock
  \url{https://github.com/vllm-project/vllm/blob/v0.31.0/vllm/v1/attention/selector.py},
  2026{\natexlab{a}}.

\bibitem[{vLLM Project}(2026{\natexlab{b}})]{vllm2026pr61192}
{vLLM Project}.
\newblock systemone: Prefill the shared prefix once before the reads on hybrid
  models.
\newblock \url{https://github.com/vllm-project/vllm/pull/61192},
  2026{\natexlab{b}}.

\bibitem[{vLLM Project}(2026{\natexlab{c}})]{vllm2026prefix}
{vLLM Project}.
\newblock Automatic prefix caching.
\newblock
  \url{https://docs.vllm.ai/en/latest/features/automatic_prefix_caching.html},
  2026{\natexlab{c}}.

\bibitem[{vLLM Project}(2026{\natexlab{d}})]{vllm2026systemone}
{vLLM Project}.
\newblock Structured decisions endpoint (/v1/systemone).
\newblock \url{https://github.com/vllm-project/vllm/pull/59299},
  2026{\natexlab{d}}.

\bibitem[{vLLM Semantic Router}(2026)]{vllmsr2026d3}
{vLLM Semantic Router}.
\newblock d3 decision models.
\newblock \url{https://huggingface.co/vllm-sr/d3}, 2026.

\bibitem[Wolf et~al.(2020)Wolf, Debut, Sanh, Chaumond, Delangue, Moi, Cistac,
  Rault, Louf, Funtowicz, Davison, Shleifer, von Platen, Ma, Jernite, Plu, Xu,
  Le~Scao, Gugger, Drame, Lhoest, and Rush]{wolf2020transformers}
Thomas Wolf, Lysandre Debut, Victor Sanh, Julien Chaumond, Clement Delangue,
  Anthony Moi, Pierric Cistac, Tim Rault, R{\'e}mi Louf, Morgan Funtowicz, Joe
  Davison, Sam Shleifer, Patrick von Platen, Clara Ma, Yacine Jernite, Julien
  Plu, Canwen Xu, Teven Le~Scao, Sylvain Gugger, Mariama Drame, Quentin Lhoest,
  and Alexander~M. Rush.
\newblock Transformers: State-of-the-art natural language processing.
\newblock In \emph{Proceedings of the 2020 Conference on Empirical Methods in
  Natural Language Processing: System Demonstrations}, pages 38--45, 2020.

\bibitem[Wu et~al.(2025)Wu, Qin, Wong, and Soatto]{wu2025stree}
Yangchao Wu, Zongyue Qin, Alex Wong, and Stefano Soatto.
\newblock {STree}: Speculative tree decoding for hybrid state space models.
\newblock In \emph{Advances in Neural Information Processing Systems}, 2025.

\bibitem[Yang et~al.(2024)Yang, Jimenez, Wettig, Lieret, Yao, Narasimhan, and
  Press]{yang2024sweagent}
John Yang, Carlos~E. Jimenez, Alexander Wettig, Kilian Lieret, Shunyu Yao,
  Karthik Narasimhan, and Ofir Press.
\newblock {SWE}-agent: Agent-computer interfaces enable automated software
  engineering.
\newblock In \emph{Advances in Neural Information Processing Systems},
  volume~37, 2024.

\bibitem[Yang et~al.(2025)Yang, Kautz, and Hatamizadeh]{yang2025gdn}
Songlin Yang, Jan Kautz, and Ali Hatamizadeh.
\newblock Gated delta networks: Improving {Mamba2} with delta rule.
\newblock In \emph{International Conference on Learning Representations}, 2025.

\bibitem[Yao et~al.(2025)Yao, Chen, Zhang, You, Yuan, Wang, and
  Lin]{yao2025deft}
Jinwei Yao, Kaiqi Chen, Kexun Zhang, Jiaxuan You, Binhang Yuan, Zeke Wang, and
  Tao Lin.
\newblock {DeFT}: Decoding with flash tree-attention for efficient
  tree-structured {LLM} inference.
\newblock In \emph{International Conference on Learning Representations}, 2025.

\bibitem[Yao et~al.(2023)Yao, Zhao, Yu, Du, Shafran, Narasimhan, and
  Cao]{yao2023react}
Shunyu Yao, Jeffrey Zhao, Dian Yu, Nan Du, Izhak Shafran, Karthik Narasimhan,
  and Yuan Cao.
\newblock {ReAct}: Synergizing reasoning and acting in language models.
\newblock In \emph{International Conference on Learning Representations}, 2023.

\bibitem[Ye et~al.(2024)Ye, Tao, Huang, and Li]{ye2024chunkattention}
Lu~Ye, Ze~Tao, Yong Huang, and Yang Li.
\newblock {ChunkAttention}: Efficient self-attention with prefix-aware {KV}
  cache and two-phase partition.
\newblock In \emph{Proceedings of the 62nd Annual Meeting of the Association
  for Computational Linguistics}, 2024.

\bibitem[Ye et~al.(2025)Ye, Chen, Lai, Lin, Zhang, Wang, Chen, Kasikci,
  et~al.]{ye2025flashinfer}
Zihao Ye, Lequn Chen, Ruihang Lai, Wuwei Lin, Yineng Zhang, Stephanie Wang,
  Tianqi Chen, Baris Kasikci, et~al.
\newblock {FlashInfer}: Efficient and customizable attention engine for {LLM}
  inference serving.
\newblock In \emph{Proceedings of Machine Learning and Systems}, 2025.

\bibitem[Zheng et~al.(2023)Zheng, Chiang, Sheng, Zhuang, Wu, Zhuang, Lin, Li,
  Li, Xing, Zhang, Gonzalez, and Stoica]{zheng2023judging}
Lianmin Zheng, Wei-Lin Chiang, Ying Sheng, Siyuan Zhuang, Zhanghao Wu, Yonghao
  Zhuang, Zi~Lin, Zhuohan Li, Dacheng Li, Eric~P. Xing, Hao Zhang, Joseph~E.
  Gonzalez, and Ion Stoica.
\newblock Judging {LLM}-as-a-judge with {MT-Bench} and {Chatbot Arena}.
\newblock In \emph{Advances in Neural Information Processing Systems, Datasets
  and Benchmarks Track}, 2023.

\bibitem[Zheng et~al.(2024{\natexlab{a}})Zheng, Yin, Xie, Sun, Huang, Yu, Cao,
  Kozyrakis, Stoica, Gonzalez, Barrett, and Sheng]{zheng2024sglang}
Lianmin Zheng, Liangsheng Yin, Zhiqiang Xie, Chuyue Sun, Jeff Huang, Cody~Hao
  Yu, Shiyi Cao, Christos Kozyrakis, Ion Stoica, Joseph~E. Gonzalez, Clark
  Barrett, and Ying Sheng.
\newblock {SGLang}: Efficient execution of structured language model programs.
\newblock In \emph{Advances in Neural Information Processing Systems},
  2024{\natexlab{a}}.

\bibitem[Zheng et~al.(2024{\natexlab{b}})Zheng, Ji, Fang, Zhou, Liu, and
  Peng]{zheng2024batchllm}
Zhen Zheng, Xin Ji, Taosong Fang, Fanghao Zhou, Chuanjie Liu, and Gang Peng.
\newblock {BatchLLM}: Optimizing large batched {LLM} inference with global
  prefix sharing and throughput-oriented token batching.
\newblock \emph{arXiv preprint arXiv:2412.03594}, 2024{\natexlab{b}}.

\bibitem[Zhu et~al.(2024)Zhu, Wang, Zhang, and Lau]{zhu2024relayattention}
Lei Zhu, Xinjiang Wang, Wayne Zhang, and Rynson W.~H. Lau.
\newblock {RelayAttention} for efficient large language model serving with long
  system prompts.
\newblock In \emph{Proceedings of the 62nd Annual Meeting of the Association
  for Computational Linguistics}, 2024.

\end{thebibliography}
